\documentclass[10.5pt]{article}

\usepackage[a4paper,margin=1in]{geometry}
\usepackage[utf8]{inputenc}
\usepackage[T1]{fontenc}
\usepackage{microtype}
\usepackage{amsmath,amssymb,amsfonts,amsthm}
\usepackage{dsfont}
\usepackage{extarrows}
\usepackage{nicefrac}
\usepackage{graphicx}
\usepackage{booktabs}
\usepackage{array,multirow,makecell,adjustbox,pifont,arydshln}
\usepackage{caption}
\usepackage[authoryear,round]{natbib}
\usepackage{xcolor}
\usepackage{hyperref}
\hypersetup{colorlinks=true,linkcolor=black,citecolor=teal,urlcolor=cyan}
\usepackage[noabbrev,capitalise]{cleveref}
\usepackage{enumitem}
\usepackage{placeins}
\usepackage{authblk}
\usepackage{mathrsfs}
\usepackage{ragged2e}
\usepackage{float}
\usepackage[most]{tcolorbox}

\tcbset{
  aibox/.style={
    width=\linewidth,
    top=7pt, bottom=2pt,
    colback=blue!6!white,
    colframe=black,
    colbacktitle=black,
    fonttitle=\bfseries\color{white},
    center title,
    enhanced,
    attach boxed title to top left={
      yshift=-0.1in, xshift=0.15in
    },
    boxed title style={
      boxrule=0pt, colframe=white,
      left=5pt, right=5pt, top=0.5pt, bottom=0.5pt
    },
  }
}
\newtcolorbox{Finding}[2][]{aibox,title={#2},#1}

\newcommand{\equalcontrib}{\textsuperscript{*}}
\newcommand{\corrauthor}{\textsuperscript{\dag}}
\newcommand{\cmark}{\textcolor{green!80!black}{\ding{51}}}
\newcommand{\xmark}{\textcolor{red!80!black}{\ding{55}}}

\newtheorem{theorem}{Theorem}[section]

\theoremstyle{definition}

\newtheorem{definition}{Definition}

\newtheorem{remark}{Remark}
\newtheorem{assumption}{Assumption}

\title{Rethinking the Unsolvable:\\ When In-Context Search Meets Test-Time Scaling}


\author[1]{\textbf{Fanzeng Xia}\equalcontrib}
\author[1]{\textbf{Yidong Luo}\equalcontrib}
\author[1]{\textbf{Tinko Sebastian Bartels}}
\author[2]{\textbf{Yaqi Xu}}
\author[1]{\textbf{Tongxin Li}\corrauthor}

\affil[1]{The Chinese University of Hong Kong, Shenzhen}
\affil[2]{Beijing University of Posts and Telecommunications}

\date{}

\begin{document}

\maketitle

\begingroup
  \renewcommand\thefootnote{\fnsymbol{footnote}}
  \footnotetext{\equalcontrib\ Equal Contribution.}
  \footnotetext{\corrauthor\ Corresponding Author.}
\endgroup

\begin{abstract}

Recent research has highlighted that Large Language Models (LLMs), even when trained to generate extended long reasoning steps, still face significant challenges on hard reasoning problems. However, much of the existing literature relies on direct prompting with simple in-context learning examples for evaluation, which largely overlooks advanced techniques to elicit LLMs' deliberate reasoning before drawing conclusions that LLMs hit a performance ceiling. In this paper, we systematically explore the combined potential of in-context search and test-time scaling on super hard reasoning tasks. We find that by employing advanced in-context search prompting to LLMs augmented with internal scaling, one can achieve transformative performance breakthroughs on tasks previously deemed ``unsolvable'' (e.g., reported success rates below 5\%). We provide both empirical results and theoretical analysis of how this combination can unleash LLM reasoning capabilities: i) Empirically, on controlled NP-Hard tasks and complex real-world planning benchmarks, our approach achieves up to a \textbf{30$\times$} improvement in success rates compared to previously reported results without any external mechanisms; ii) Theoretically, we show that in-context search prompting, when combined with internal scaling, significantly extends the complexity class of solvable reasoning problems. These findings challenge prevailing assumptions about the limitations of LLMs on complex tasks, indicating that current evaluation paradigms systematically underestimate their true potential. Our work calls for a critical reassessment of how LLM reasoning is benchmarked and a more robust evaluation strategy that fully captures the true capabilities of contemporary LLMs, which can lead to a better understanding of their operational reasoning boundaries in real-world deployments.

\end{abstract}

\section{Introduction}

Large Language Models (LLMs) have demonstrated significant progress in solving diverse reasoning and planning tasks, such as instruction following~\citep{wei2021finetuned, ouyang2022training}, commonsense reasoning~\citep{talmor2018commonsenseqa, brown2020language}, and code generation~\citep{chen2021evaluating, austin2021program}.
Despite their achievements, LLMs still face significant challenges when applied to complex, long-horizon problem instances. The emergence of reasoning-oriented models, as exemplified by~\citep{jaech2024openai, guo2025deepseek}, has led to substantial progress in solving problems that require extensive reasoning and planning. Trained through reinforcement learning and high-quality trajectories to develop systematic reasoning processes, these models demonstrate improved performance on tasks that require multi-step deduction and self-correction. This enhanced capability, evidenced by benchmarks such as PlanBench~\citep{valmeekam2023planbench}, marks a significant advancement in areas where earlier LLMs exhibited notable deficiencies. However, this improvement is not uniform across all problem complexities; the models particularly struggle with more challenging instances, such as controlled NP-hard tasks like Vertex Cover and 3-Dimensional Matching~\citep{yang2025nondeterministic}, and sophisticated real-world planning scenarios such as the Natural Plan Benchmark~\citep{zheng2024natural}.
Recent critiques~\citep{valmeekam2024llms, valmeekam2024planning} emphasize that difficulties persist in solving these complex problem instances, highlighting the need for further strategies to address the full spectrum of reasoning challenges in real-world applications.

To optimize the deployment of LLMs across diverse problem difficulties, recent studies~\citep{chen2024unlocking, sun2025climbing} have sought to define their reasoning limitations. 
A common observation is that current LLMs encounter a significant performance ceiling (often when success rates on complex tasks drop below $10$-$20\%$), suggesting that additional training with new architectures may be needed for breakthroughs on hard problems. However, our analysis of experimental settings in recent studies (see~\Cref{table:related_work} in Appendix~\ref{app:related_work}) reveals a pattern: most evaluations rely on \textbf{direct prompting} strategies. These include few-shot in-context learning without task decomposition~\citep{wang2020generalizing}, automatic Chain-of-Thought (Auto-CoT) prompting~\citep{zhang2022automatic} where models generate reasoning steps autonomously, and hints before prompting~\citep{fu2024hint}. 
Focusing mostly on direct prompting, prior work might have missed the power of advanced in-context search prompting techniques to elicit deeper reasoning, possibly leading to a biased assessment of LLMs' actual reasoning boundaries. Inspired by this, our work aims to systematically probe and potentially elevate the perceived upper limits of LLM reasoning on ``super hard'' problem instances (those with previously reported success rates below 5\%). This investigation motivates the following central question: 

 \begin{figure}[t!]
    \centering    
    \hspace{-8mm}
    \includegraphics[width=145mm]{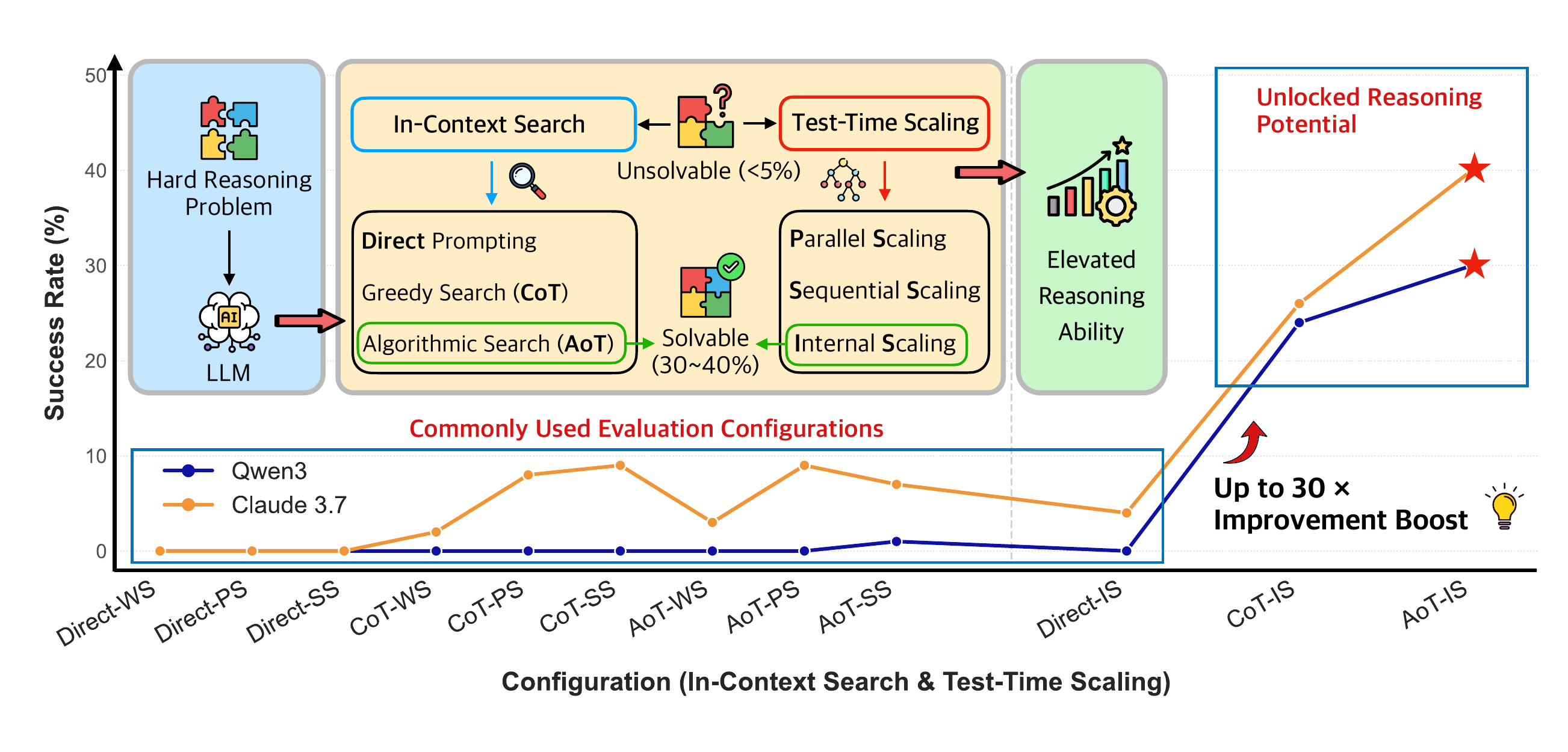}
    \caption{Overview of promoted LLM reasoning boundaries for hard problems via the combination of in-context search and test-time scaling. Our findings reveal that tasks previously reported as \textit{unsolvable} are actually \textit{solvable} and thus require a \textbf{rethink} of the current evaluation configurations. Evaluation is based on the \textit{Trip Planning} task as an illustrative example.}
    \label{fig:intro}
    \vspace{-2mm}
\end{figure}

\begin{center}
\textit{Are LLMs truly incapable of solving tasks deemed ``unsolvable'' by current evaluation paradigms, or can their full potential on such problems yet be unleashed?}
\end{center}

To answer this question, we adopt a standard prompt-based approach (without external mechanisms or additional fine-tuning) to evaluate the upper-bound reasoning performance of out-of-the-box LLMs on two challenging problem categories: controlled NP-hard tasks and complex real-world planning scenarios. Overall, there are two types of methods commonly employed to enhance the on-the-fly reasoning capabilities of LLMs: (i) \textbf{In-context search prompting}, a class of prompting strategies that enables LLMs to internally simulate search processes through learned in-context examples. Our experiments explore a variety of prompting techniques designed to optimize model performance across diverse tasks, including direct prompting, Chain-of-Thought (CoT), and Algorithm-of-Thought (AoT); (ii) \textbf{Test-time scaling}, a class of techniques that augment LLM reasoning at inference time by expanding computational effort or diversifying reasoning pathways. We experiment with methods including parallel scaling, sequential scaling, and internal scaling. Further details on both these categories of methods are provided in Appendix~\ref{app:related_work}.

Recent representative works have sought to provide insights into improving LLM reasoning capabilities from different perspectives: (i) Through in-context search prompting,~\citep{ge2025innate} explored how different direct prompting strategies affect models with internal scaling, while~\citep{sel2025llms} developed advanced in-context search prompting methods that enable LLMs to better follow structured in-context search processes; (ii) Regarding test-time scaling,~\citep{snell2024scaling} investigated the impact of parallel scaling and sequential scaling with direct prompting, leaving more advanced in-context search techniques unexplored, concluding that these test-time scaling schemes yielded only small gains on hard problems; (iii) In the context of super-hard tasks,~\citep{chen2024reprompt} examined sequential scaling in conjunction with external pipelines for complex real-world planning. Other investigations into NP-hard problems, such as 3-SAT and Sum-of-Squares~\citep{hazra2025have, li2025sos1}, have found that LLMs show promise on NP-hard instances and have probed the models to understand different cognitive behaviors. These studies provide initial explorations of specific NP-hard tasks, but an overall understanding and analysis are still lacking.

While these prior works offer valuable insights, we identify a crucial gap in the current literature as shown in~\Cref{table:related_work}: the potential of advanced in-context search methods, particularly when deeply integrated with test-time scaling strategies, remains unexplored. Specifically, how their fine-grained combinations might advance the current understanding and quantification of the true reasoning capacity of LLMs on unsolvable tasks is still a critical open problem. As illustrated in~\Cref{fig:intro}, our ablation studies reveal a significant empirical finding: by combining advanced in-context search prompting as algorithmic guidance for LLMs endowed with internal scaling, we achieve \textbf{up to 30$\times$ improvement} on problem instances previously considered unsolvable and intractable (reported success rates below 5\%). These results indicate that commonly employed configurations for evaluating the true reasoning ability of contemporary LLMs systematically underestimate the attainable reasoning ceiling, thereby masking their operational boundaries in real deployments.

To further investigate the underlying reasons behind these improvements, we extend prior theoretical work in~\citep{merrill2023expressive, sel2023algorithm}, providing an analysis of the expressivity of LLMs augmented with in-context search and internal scaling. Collectively, our empirical and theoretical results necessitate a systematic reassessment of current methodologies for probing LLM reasoning boundaries and call for the development of more robust and faithful evaluation techniques.

Our main contributions are summarized as follows: 
\begin{itemize}[leftmargin = 6mm] 
    \item \textbf{Identifying Systematic Bias.} We reveal systematic evaluation bias in the commonly used evaluation configurations of existing research that underestimates the attainable performance ceiling of LLMs' reasoning ability and can yield misleading operational boundaries. 
    \item \textbf{Improved LLM Reasoning Abilities.} By conducting ablation studies on fine-grained combinations of in-context search prompting with test-time scaling, we achieve up to a 30× improvement compared to the previously reported success rates on two controlled NP-hard tasks and two complex real-world planning problems.
    \item \textbf{Theoretical Insights.} We provide a theoretical analysis of the reasoning boundary that can be achieved with the combination of in-context search and internal scaling, offering valuable insights into the expressive power of contemporary LLMs. 
\end{itemize}

\vspace{-5pt}

\section{Empirical Results}
\label{sec:exp}

\vspace{-5pt}


In this section, we present the empirical results supporting our central claim. We begin by describing the setup design of our experiments on two hard reasoning problem instances. Following this, we present our findings through a detailed three-level ablation study to examine combinations of in-context search and test-time scaling methods to identify systematic bias in commonly used evaluation configurations.

\subsection{Experimental Setup}
\label{sec:setup}

\textbf{Tasks.} Our investigation incorporates two categories of super hard task instances: (i) \textbf{Controlled NP-Hard Tasks}: we begin by conducting experiments on controlled Nondeterministic Polynomial-time (NP)-Hard problems, which are problems requiring exponential time to find solutions. Specifically, we select Vertex Cover and 3-Dimensional Matching (3DM) and generate 100 controlled problem instances according to the methods in~\citep{yang2025nondeterministic}, enabling an understanding of their average behaviors and reliable performance estimates. We evaluate the LLMs' reasoning capabilities on tasks at the most challenging difficulty (level 10). (ii) \textbf{Complex Real-World Planning}: furthermore, we disentangle the 
\begin{table}[H]
\centering
\small 
\caption{Performance on \textit{Vertex Cover} (Difficulty Level = 10) for controlled NP-hard task. Qwen3 showed almost no improvement across all methods, possibly because of struggling with complex numerical abstract reasoning.}
\label{table:vertex_cover} 

\begin{tabular*}{\textwidth}{@{}p{1.8cm} 
                             @{\extracolsep{\fill}} 
                             >{\RaggedRight\arraybackslash}p{5.0cm} 
                             >{\centering\arraybackslash}p{2.4cm} 
                             @{\extracolsep{\fill}} 
                             >{\centering\arraybackslash}p{2.4cm} 
                             @{\extracolsep{\fill}} 
                             >{\centering\arraybackslash}p{2.6cm}@{}} 
\toprule
\multirow{2}{*}{\textbf{Model}} & \multirow{2}{*}{\textbf{Evaluation Strategy}} & \multicolumn{1}{c}{Direct} & \multicolumn{1}{c}{Greedy Search} & \multicolumn{1}{c}{Depth-First} \\
& & \multicolumn{1}{c}{Prompting} & \multicolumn{1}{c}{(CoT)} & \multicolumn{1}{c}{Search (AoT)} \\
\midrule
\multirow{4}{*}{\textbf{Qwen3}} 
& No Scaling (Base Model) & 0/100 = 0\% & 0/100 = 0\% & 0/100 = 0\% \\
& Parallel Scaling (Best-of-N) & 0/100 = 0\% & 0/100 = 0\% & 0/100 = 0\% \\
& Sequential Scaling (Self-Refine) & 0/100 = 0\% & 0/100 = 0\% & 0/100 = 0\% \\
& Internal Scaling (Thinking Mode) & 0/100 = 0\% & 1/100 = 1\% & 2/100 = 2\% \\
\midrule
\multirow{4}{*}{\textbf{Claude 3.7}} 
& No Scaling (Base Model) & 0/100 = 0\% & 1/100 = 1\% & 1/100 = 1\% \\
& Parallel Scaling (Best-of-N) & 0/100 = 0\% & 4/100 = 4\% & 4/100 = 4\% \\
& Sequential Scaling (Self-Refine) & 0/100 = 0\% & 5/100 = 5\% & 3/100 = 3\% \\
& Internal Scaling (Thinking Mode) & 2/100 = 2\% & 21/100 = \textbf{21\%} & 31/100 = \textbf{31\%} \\
\bottomrule
\end{tabular*}
\end{table}

\begin{table}[htbp!]
\centering
\small 
\caption{Performance on \textit{Trip Planning} (Difficulty Level = 10) for complex real-world planning. When the complexity of numerical inputs in language-based planning is disentangled, Qwen3 improved.}
\label{table:trip_planning} 

\begin{tabular*}{\textwidth}{@{}p{1.8cm} 
                             @{\extracolsep{\fill}} 
                             >{\RaggedRight\arraybackslash}p{5.0cm} 
                             >{\centering\arraybackslash}p{2.4cm} 
                             @{\extracolsep{\fill}} 
                             >{\centering\arraybackslash}p{2.4cm} 
                             @{\extracolsep{\fill}} 
                             >{\centering\arraybackslash}p{2.6cm}@{}} 
\toprule
\multirow{2}{*}{\textbf{Model}} & \multirow{2}{*}{\textbf{Evaluation Strategy}} & \multicolumn{1}{c}{Direct} & \multicolumn{1}{c}{Greedy Search} & \multicolumn{1}{c}{Depth-First} \\ 
& & \multicolumn{1}{c}{Prompting} & \multicolumn{1}{c}{(CoT)} & \multicolumn{1}{c}{Search (AoT)} \\ 
\midrule
\multirow{4}{*}{\textbf{Qwen3}} 
& No Scaling (Base Model) & 0/100 = 0\% & 0/100 = 0\% & 0/100 = 0\% \\
& Parallel Scaling (Best-of-N) & 0/100 = 0\% & 0/100 = 0\% & 0/100 = 0\% \\
& Sequential Scaling (Self-Refine) & 0/100 = 0\% & 0/100 = 0\% & 1/100 = 1\% \\
& Internal Scaling (Thinking Mode) & 0/100 = 0\% & 24/100 = \textbf{24\%} & 30/100 = \textbf{30\%} \\
\midrule
\multirow{4}{*}{\textbf{Claude 3.7}} 
& No Scaling (Base Model) & 0/100 = 0\% & 2/100 = 2\% & 3/100 = 3\% \\
& Parallel Scaling (Best-of-N) & 0/100 = 0\% & 8/100 = 8\% & 9/100 = 9\% \\
& Sequential Scaling (Self-Refine) & 0/100 = 0\% & 9/100 = 9\% & 7/100 = 7\% \\
& Internal Scaling (Thinking Mode) & 4/100 = 4\% & 26/100 = \textbf{26\%} & 40/100 = \textbf{40\%} \\
\bottomrule
\end{tabular*}
\end{table}
complexity of numerical inputs and evaluate LLMs while planning in natural language using the Trip Planning and Meeting Planning tasks from the Natural Plan benchmark~\citep{zheng2024natural}. Similarly, we randomly select 100 instances for each of these tasks. This type of task can essentially be understood as a natural language-based variation of the NP-Hard Traveling Salesperson Problem (TSP). We also select the most difficult level (level 10) for these tasks to explore the performance ceiling of LLM reasoning capabilities.

\textbf{Models.} We use an open-source LLM (Qwen3 235B~\citep{qwenlm_qwen_3} from Alibaba Cloud) and a closed-source API (Anthropic's Claude 3.7 Sonnet~\citep{anthropic_claude_3.7_sonnet}) in our experiments. These particular models were chosen because they offer functionalities that allow for the explicit activation or deactivation of an internal ``thinking'' mode. This feature is crucial as it enables us to perform ablation studies on the same base model, thereby isolating the effects of using internal scaling versus not using it. All prompting strategies were evaluated using these two models to ensure a comprehensive understanding of their capabilities and responses to different experimental conditions.

\textbf{Prompts.} We investigate three in-context search prompting strategies to instruct the LLMs for the tasks outlined above:
\begin{itemize}[leftmargin = 4mm]
    \item \textbf{Direct Prompting}: this approach provides the model with a few problem-solution pairs~\citep{wang2020generalizing} without any explicit intermediate reasoning steps in the examples. It relies on the model’s internalized reasoning to bridge from the problem to the solution and generally does not delve into deep task decomposition.
    \item \textbf{CoT Prompting}: CoT~\citep{wei2022chain} guides LLMs to generate a sequence of intermediate reasoning steps before arriving at a final solution by providing an in-context example that demonstrates a successive, step-by-step solution path (typically representing a \textit{greedy search} without algorithmic operations like backtracking).
    \item \textbf{AoT Prompting}: this approach~\citep{sel2025llms} guides the model through explicit, high-level \textit{algorithmic search} pathways. It uses detailed examples to demonstrate structured algorithmic operations, helping the model simulate decision-making processes based on algorithmic principles. 
\end{itemize}

The prompts constructed are also evaluated under 3 different test-time scaling methods: 
\begin{itemize}[leftmargin = 4mm]
    \item \textbf{Parallel Scaling}: this method improves reasoning performance by generating multiple outputs in parallel and then aggregating them. In our experiments, we specifically employed a Best-of-N (BoN) approach where N=3; the highest accuracy was recorded. 
    \item \textbf{Sequential Scaling}: this is an iterative method where subsequent computations are explicitly directed based on the results of the previous one, allowing the model to refine solutions round by round. We implemented this using the Self-Refine technique~\citep{madaan2023self}, where initial responses were augmented with Self-Refine prompts and resubmitted for improvement.
    \item \textbf{Internal Scaling}: this approach relies on the model autonomously determining the computational effort to allocate for reasoning using its internal parameters and a learned policy. For our experiments, this involved activating the ``thinking'' mode for extended reasoning paths of the selected base models.
\end{itemize}

Templates for these prompt strategies can be found in the supplementary Appendix~\ref{app:setup}, with more introduction delegated to Related Work in Appendix~\ref{app:related_work}.

\textbf{Metrics.} We measure the reasoning performance using Success Rate (referred to as Accuracy in~\cite{yang2025nondeterministic} and Solve Rate in~\cite{zheng2024natural}), defined as the percentage of problem instances for which the LLM generates a complete and verifiably correct solution within each problem domain. For the controlled NP-Hard tasks, a success is a solution that satisfies all defined problem constraints, while for the complex real-world planning tasks, it requires the generated plan to achieve an exact match with the ground truth or optimal solution. 

\subsection{Ablation Studies}

In this section, we present the results of a three-level ablation study focusing on the hard problem instances described in Section~\ref{sec:setup}. Detailed empirical results are provided in Table~\ref{table:vertex_cover},~\ref{table:trip_planning},~\ref{table:3dm}, and~\ref{table:meeting_planning}. Our discussion draws upon these tables, using the Trip Planning task (trends illustrated in Figure~\ref{fig:intro}) as an illustrative example. Results on the other tasks are quantitatively similar, with details deferred to Appendix~\ref{app:experiment}. We aim to progressively improve the reasoning boundary of LLMs and identify systematic bias in commonly adopted evaluation configurations.

\textbf{Level 1: Test-time Scaling.}
We begin with the most commonly used evaluation configuration: direct prompting augmented with four test-time scaling variants: without scaling (Direct-WS), parallel scaling (Direct-PS), sequential scaling (Direct-SS), and internal scaling (Direct-IS). Figure~\ref{fig:intro} shows that: (i) Both Qwen 3 and Claude 3.7 achieve a \mbox{0\%} success rate under Direct-WS/PS/SS. This echoes reported results from~\citep{valmeekam2024llms, chen2024reprompt, lee2025evolving}, confirming that this instance is effectively unsolvable for basic LLMs; (ii) When the model is endowed with internal scaling capabilities by switching from a ``no-thinking'' to a ``thinking'' mode, Claude 3.7 improves marginally to \mbox{4\%}, whereas Qwen 3 remains at \mbox{0\%}. This aligns with the common understanding that internal scaling can raise the upper bound of a model’s reasoning ability. However, the success rate is still < 5\%. Related literature evaluating the reasoning capabilities of state-of-the-art LLMs often stops at this point, concluding that LLM reasoning cannot solve such difficult tasks.

\begin{Finding}{Takeaway 1}
Under direct prompting (which includes methods like Auto-CoT, zero-shot, few-shot, and hints), LLMs generally struggle with hard reasoning tasks. Enabling test-time scaling yields only marginal gains and does not resolve the unsolved instances. This aligns with the conclusions from most existing research evaluations, indicating that the introduction of state-of-the-art test-time scaling methods is still insufficient to solve these hard problems.
\end{Finding}

\vspace{2mm}

\textbf{Level 2: In-Context Search.} The second level of our ablation study explores the efficacy of advanced in-context search prompting, specifically employing CoT prompting and AoT prompting. Both methods require an understanding of the task's solution path and correspondingly designed in-context examples. We evaluate these methods without test-time scaling. As Figure~\ref{fig:intro} illustrates: (i) under both CoT‑WS and AoT‑WS, Claude 3.7’s success rate rises by 2\% and 3\% respectively, whereas Qwen 3 remains fixed at \mbox{0\%}. This indicates that advanced in‑context search prompting alone can provide a performance lift on hard reasoning tasks; (ii) Along with the results from Level 1, we find that Qwen 3 consistently stays at \mbox{0\%} when either test-time scaling or advanced in-context search is applied in isolation. In contrast, Claude 3.7 shows marginal gains in both scenarios. This suggests that the effectiveness of all in-context search and test-time scaling variants is highly correlated with the inherent capabilities of the base model.

\begin{Finding}{Takeaway 2} 
Under advanced in-context search prompting with task decomposition such as CoT and AoT, LLMs exhibit improved performance on hard reasoning tasks, even without any test-time scaling. However, their standalone impact on difficult tasks is still marginal. 
\end{Finding}

\vspace{2mm}

\textbf{Level 3: Combining In-Context Search and Test-time Scaling.} The final level explores the upper-bound reasoning capabilities achieved by combining In-Context Search and Test-time Scaling across six different configurations. As illustrated in Figure~\ref{fig:intro}, these include CoT with parallel scaling (CoT-PS), CoT with sequential scaling (CoT-SS), CoT with internal scaling (CoT-IS), AoT with parallel scaling (AoT-PS), AoT with sequential scaling (AoT-SS), and AoT with internal scaling (AoT-IS). The results reveal the following key findings: (i) Under CoT-PS/SS and AoT-PS/SS, Qwen 3 for the first time shows a marginal improvement of only 1\% on AoT-SS, while Claude 3.7 demonstrates improvements of 7-9\%; (ii) The most significant breakthroughs occur with CoT-IS and AoT-IS with internal scaling. We observe substantial improvements: Qwen 3's success rate jumps to 24\% (CoT-IS) and 30\% (AoT-IS), while Claude 3.7 reaches 26\% (CoT-IS) and 40\% (AoT-IS). This represents up to a 30-fold improvement in success rate compared to the previous commonly used configurations. These results demonstrate a significant improvement in reasoning performance, breaking the previously observed evaluation thresholds for LLMs on this benchmark even when no external mechanisms were utilized. This suggests that current evaluation methods do not fully unleash the reasoning potential of LLMs, and that the current understanding of their operational boundaries is subject to systematic bias.

\vspace{2mm}
\begin{Finding}{Takeaway 3} When combining in-context search prompting with test-time scaling, especially CoT/AoT with internal scaling, LLMs achieve up to a \textbf{$30\times$} improvement in success rates on hard reasoning tasks. This reveals the significant untapped potential of LLMs that commonly used evaluation methods fail to capture, highlighting a call for more robust and faithful evaluation techniques to define LLMs' true operational boundaries in real deployments. \end{Finding}

\vspace{-5pt}

\section{Theoretical Analysis} \label{theory}
\vspace{-5pt}

Driven by our empirical results, we explore the theoretical foundations of why combining in-context search methods (CoT and AoT) with internal scaling enables LLMs to solve tasks that were previously unsolvable and intractable. We aim to establish a formal basis for understanding how these methods can redefine and extend the perceived reasoning boundaries of LLMs.
\vspace{-3pt}
\subsection{Preliminaries}
\vspace{-3pt}
To lay the groundwork for our theoretical analysis, we first define the key in-context search prompting strategies that are central to our investigation and subsequent theorems.

\begin{definition}[\emph{In-Context Search Prompting}] 

The following defines the primary in-context search methods employed and analyzed in this work, progressing from simpler to more structured forms of reasoning guidance:
\begin{itemize}[leftmargin=4mm]

    \item \textbf{Direct Prompting (Few-Shot Learning):}  The model is augmented only with a set of a few Problem–Solution pairs, without any intermediate reasoning steps. Concretely, given a target input (prompt) $x=(x_1, \dots, x_n)$, $m$ in-context examples are prepended. These examples form the demonstration sequence $S_{DP}$:

    \[ S_{DP} = \left(x^{(1)},y^{(1)},\,x^{(2)},y^{(2)},\,\ldots,\,x^{(m)},y^{(m)}\right), \]

    where each $x^{(j)}$ (for $j=1, \dots, m$) is an example problem instance and each $y^{(j)}$ is its corresponding solution. Note that the target input $x$ is a distinct problem instance from the example inputs $x^{(j)}$ provided in $S_{DP}$. The model then emits its answer $y_{ans}$ for the target input $x$ conditioned on the full sequence formed by $S_{DP}$. Direct prompting provides only question-answer supervision in each example pair, relying on the model’s internalized reasoning to bridge from problem to solution without explicit in-context search guidance.

\item \textbf{Chain of Thought (CoT):} CoT is a greedy in-context search paradigm wherein a transformer model, upon receiving an input $x=(x_1, \dots, x_n)$ and a \textit{successive} in-context search example (without algorithmic steps, e.g. backtracking) 
\[
  \bigl[\,\mathrm{Problem},\;\mathsf{Greedy\ Search\ Trace},\;\mathrm{Solution}\bigr]
\] 
that demonstrates step-by-step reasoning, generates a sequence of intermediate tokens \(S_{CoT} = (s_1, s_2, \dots, s_{t(n)})\) of up to \(t(n)\) auxiliary tokens through its autoregressive decoding process. Before giving a final answer token \(y_{ans}\), each intermediate token \(s_j\) is produced conditioned on the input and all previously generated tokens:
  \[
    s_j \;=\;\mathrm{Transformer}_\theta\bigl(x,\;s_1,\dots,s_{j-1}\bigr),
    \quad j=1,\dots,t(n).
  \]
  The class of languages recognizable by a Transformer that uses at most \(t(n)\) such intermediate decoding steps (i.e., generates up to \(t(n)\) intermediate tokens) is denoted \(\mathsf{CoT}(t(n))\). In essence, CoT allows the model to use these generated intermediate tokens as a form of working memory or recurrent state to enable sequential reasoning.

\item \textbf{Algorithm of Thought (AoT):} AoT further extends CoT by providing algorithmic in-context search examples mimicking an explicit search algorithm.  Upon processing an input $x=(x_1, \dots, x_n)$, a transformer model is elicited by an \textit{algorithmic} in-context search example
\[
  \bigl[\,\mathrm{Problem},\;\mathsf{Algorithmic\ Search\ Trace},\;\mathrm{Solution}\bigr]
\]
to generate a sequence of intermediate tokens \(S_{AoT} = (s_1, s_2, \dots, s_{a(n)})\) of up to \(a(n)\) auxiliary tokens. 
These tokens collectively instantiate an algorithmic search trace for the given problem, following the same underlying autoregressive augmentation mechanism as CoT. This paradigm empowers the model to generate tokens representing distinct algorithmic operations, including but not limited to:
  \begin{enumerate}
    \item \textsc{Initialization:} Defining the search space or partitioning the problem into subproblems;
    \item \textsc{Expansion:} Traversing a  path or branch within the search tree (e.g., depth-first exploration);  
    \item \textsc{Evaluation:} Assessing the branch’s promise (prune if needed);
    \item \textsc{Backtracking:} Reverting to a previous node to explore alternative branches or strategies.
  \end{enumerate}
  We denote by \(\mathsf{AoT}(a(n))\) the class of languages the transformer can recognize when augmented by generating \(a(n)\) reasoning tokens. By structuring the prompt as such, AoT unlocks the model’s ability to carry out complex, multi-path reasoning in a single query without relying on external search mechanisms.

\end{itemize}

\vspace{2mm}

\end{definition}
\begin{definition}[\emph{Internal Scaling}]

Internal scaling refers to a mechanism that enables a model to autonomously scale the amount of intermediate tokens to allocate for a given problem at test time. This determination is guided by the model’s learned parameters $\theta$ and an internal control policy $\pi_{\theta}$, acquired from reasoning-oriented training~\citep{anthropic_claude_3.7_sonnet, qwenlm_qwen_3}. 

The total number of generated tokens or reasoning steps $T$ is thus dynamically determined by this internally trained mechanism rather than being preset. The halting of the process for deciding a language is ensured by the functioning of the learned policy $\pi_{\theta}$ to stop within a finite $T$. By adapting its computational effort based on the perceived requirements of the input, it can lead to emergent test-time behaviors such as generating more detailed and extended reasoning chains for complex problems, or performing self-correction and evaluation steps during the process. In our work, we switch between thinking and no-thinking modes to simulate with or without internal scaling setting: 
\begin{itemize}[leftmargin=4mm]
\setlength\itemsep{-0.3em}
    \item Without internal scaling (no thinking mode), the policy $\pi_{\theta}$ will lead to a small $T$, which we assume will be bounded by a polynomial function of the input size $n$ (i.e., $T = \mathtt{poly}(n)$).
    \item With internal scaling (thinking mode), the policy $\pi_{\theta}$ when faced with a task requiring deeper reasoning, can allow $T$ to scale significantly further, potentially to an exponential function of $n$ (i.e., $T = \mathtt{exp}(n)$).
\end{itemize}
The dynamically allocated number of reasoning steps $T$ directly corresponds to the length of the generated sequence of thought tokens (such as $t(n)$ in the definition of $\mathsf{CoT}(t(n))$ or $a(n)$ in $\mathsf{AoT}(a(n))$). In the next section, we show that the ability of internal scaling to modulate $T$ from polynomial to exponential scales is what allows such models to potentially address correspondingly more complex problems. 

\end{definition}

Further theoretical definitions, including Turing machine, computational complexity classes, and decoder-only transformers are provided in Appendix~\ref{app:theory}.

\subsection{Theoretical Results}

Standard Transformer models, characterized by a fixed number of layers, face inherent constraints on the depth of sequential computation they can natively perform. Without augmentation, their expressive power is typically confined to complexity classes significantly weaker than $\mathsf{P}$~\citep{chen2024theoretical}. CoT prompting has revolutionized the approach to complex reasoning tasks with LLMs. This technique guides the model to generate an explicit sequence of intermediate reasoning steps before producing a final output. This explicit articulation of reasoning allows the model to undertake multi-step computations that are more complex than would be possible with direct-answer generation.
\begin{figure}[h]
    \centering    
\includegraphics[width=125mm]{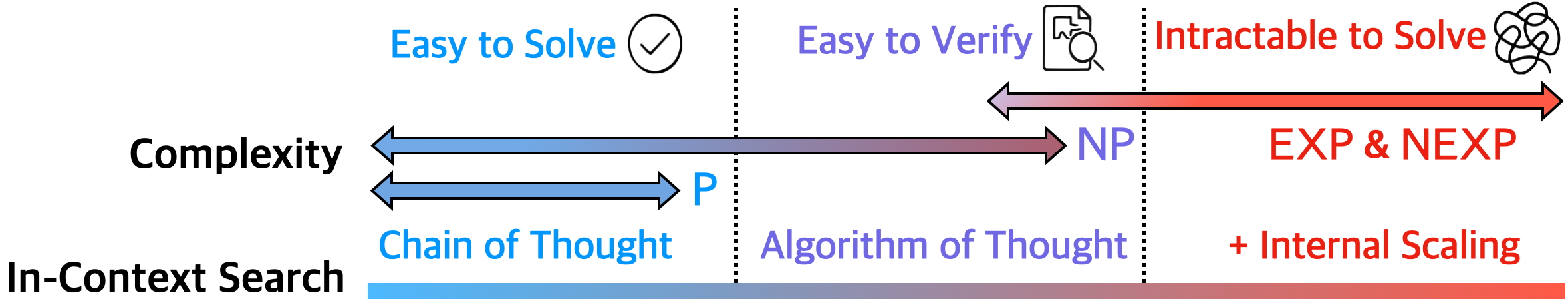}
    \vspace{1mm}
    \caption{Conceptual roadmap illustrating the power of in-context search and test-time scaling in pushing the reasoning boundary of LLMs. LLMs can solve problems in $\mathsf{P}$ (Theorem \ref{thm:CoT}) and $\mathsf{NP}$ (Theorem \ref{thm:AoT}) using standard CoT and AoT with polynomial-length traces.  Internal Scaling, by extending these thought processes to exponential lengths, significantly pushes the reasoning boundary towards $\mathsf{EXP}$ (Theorem \ref{thm:CoT_exp}) and $\mathsf{NEXP}$ (Theorem \ref{thm:AoT_exp}).}
    \label{fig:theory}
    \vspace{-1mm}
\end{figure}

Drawing upon recent theoretical advances~\citep{li2024chain, merrill2023expressive}, we begin with the established understanding that Transformers using a polynomial number of CoT steps are computationally equivalent to polynomial-time Deterministic Turing Machines (DTMs), enabling them to solve problems in the class $\mathsf{P}$. Figure~\ref{fig:theory} offers a conceptual roadmap that illustrates how the subsequent theorems progressively extend Transformers' reasoning boundary to higher complexity classes with different configurations. We first recall the following theorem  from \citep{merrill2023expressive}:

\begin{theorem}[\emph{$\mathsf{CoT}(\mathtt{poly}(n)) = \mathsf{P}$}~\cite{merrill2023expressive}] 
\label{thm:CoT}

The class of languages decidable by decoder-only transformer models (satisfying the architectural assumptions in Assumption~\ref{def:transformer} detailed in the supplementary appendix), when augmented with a CoT of length $t(n) = \mathtt{poly}(n)$, is the complexity class $\mathsf{P}$. 
\end{theorem}

\vspace{1mm}

Therefore, Transformers equipped with a polynomial number of CoT steps are computationally equivalent to polynomial-time DTMs. Building on this, when Transformers are prompted with AoT to leverage their capacity for simulating non-deterministic search processes, their computational power extends to the class NP for polynomial-length reasoning traces.
\vspace{0.5mm}

\begin{theorem}[\emph{$\mathsf{AoT}(\mathtt{poly}(n)) = \mathsf{NP}$}] 
\label{thm:AoT}
The class of languages decidable by decoder-only Transformer models (satisfying the architectural assumptions in Assumption~\ref{def:transformer}), when prompted with AoT such that the total number of intermediate tokens generated along any single accepting computational path is bounded by a polynomial $a(n) = \mathtt{poly}(n)$, is the complexity class NP. 

\end{theorem}

The proof of Theorem~\ref{thm:AoT} is given in Appendix~\ref{subsec:proof-AoT}. The length of the generated thought sequence is a crucial factor. If the number of Chain of Thought steps is allowed to scale exponentially with the input size, Transformers can solve problems from correspondingly higher complexity classes. This exponential scaling might initially seem impractical due to computational demands, but for practical instances of hard problems as demonstrated in~\Cref{sec:exp}, it remains entirely feasible. The proof is given in Appendix~\ref{subsec:proof-CoT_exp}. 

\vspace{0.5mm}

\begin{theorem}[\emph{$\mathsf{CoT}(\mathtt{exp}(n)) = \mathsf{EXP}$}] 
\label{thm:CoT_exp}
The class of languages decidable by decoder-only transformer models (satisfying the architectural assumptions in Assumption~\ref{def:transformer}), when augmented with a Chain of Thought (CoT) of length $t(n) = \mathtt{exp}(n)$ (i.e., $t(n) = O(2^{p(n)})$ for some polynomial $p(n)$ in the input length $n$), is the complexity class $\mathsf{EXP}$. 
\end{theorem}
\vspace{1mm}

Similarly, this exponential scaling of reasoning steps also enhances the power of AoT, enabling Transformers to tackle problems within non-deterministic exponential time. The proof is given in Appendix~\ref{subsec:proof-AoT_exp}. 
\vspace{0.5mm}

\begin{theorem}[\emph{$\mathsf{AoT}(\mathtt{exp}(n)) = \mathsf{NEXP}$}] 
\label{thm:AoT_exp}
The class of languages decidable by decoder-only Transformer models (satisfying the architectural assumptions in Assumption~\ref{def:transformer}), when prompted with Algorithm of Thought (AoT) such that the total number of intermediate tokens generated along any single accepting computational path is bounded by an exponential function $a(n)=\mathtt{exp}(n)$ (i.e., $a(n)=O(2^{p_A(n)})$ for some polynomial $p_A(n)$ in the input length $n$), is the complexity class $\mathsf{NEXP}$. 
\end{theorem}

\vspace{1mm}

While these theorems establish a clear link between the length of generated thought processes and computational power, it is important to emphasize that the mere quantity of tokens is not the sole determinant. The following theorem highlights the critical role of computationally relevant core reasoning steps:

\vspace{0.5mm}

\begin{theorem}[{Core Reasoning Tokens}] 
\label{thm:core_reasoning_tokens}
For a decoder-only Transformer model (satisfying Assumption~\ref{def:transformer}), the decidable complexity class is predicated on the length of its \emph{core computational trace}, $k_{\text{core}}(n)$, consisting of only computationally essential steps; generating additional redundant tokens (increasing total tokens $k_{\text{total}}(n)$ beyond $k_{\text{core}}(n)$) does not elevate this class. Specifically, if $k_{\text{core}}(n) = O(\mathtt{poly}(n))$, the model decides languages within $\mathsf{P}$ (if its internal algorithm $A$ is deterministic) or $\mathsf{NP}$ (if $A$ is non-deterministic). Consequently, to decide languages in $\mathsf{EXP}$ or $\mathsf{NEXP}$ respectively, a core computational trace of $k_{\text{core}}(n) = O(\mathtt{exp}(n))$ is necessary.
\end{theorem}

Theorem~\ref{thm:core_reasoning_tokens} aligns with the recent empirical observations that the quality and computational relevance of generated thoughts are more critical than the length of reasoning tokens for effective complex reasoning~\citep{zeng2025revisiting, li2025sos1, wu2025more, ballon2025relationship, su2025between}. The proof is given in Appendix~\ref{subsec:proof-core_reasoning_tokens}.

\section{Conclusion}

Our results demonstrate that LLMs can significantly outperform previous expectations on challenging tasks, such as NP-hard problems and complex real-world planning, achieving success rates up to 30 times higher through advanced in-context search and test-time scaling techniques. These findings together with the theoretical evidence challenge the assumption that LLMs have reached their performance limits without external mechanisms, highlighting the need for improved evaluation methods to fully uncover and assess LLMs' true reasoning capabilities.

\textbf{Limitations and Future Directions.} While this work reveals substantial untapped LLM reasoning potential, future efforts should focus on developing more efficient, automated reasoning strategies and deepening our mechanistic understanding to fully realize these extended capabilities for broader deployment: i) extending the theoretical analysis, which currently emphasizes internal scaling, to comprehensively cover parallel and sequential scaling methods; ii) diversifying the in-context search algorithm examples beyond Depth-First Search to include techniques like Monte Carlo Tree Search and Graph Search; iii) developing more efficient and automated reasoning strategies, for instance, to reduce the current reliance on potentially well-defined, hand-crafted prompts or examples; iv) exploring the potential of hybrid test-time scaling approaches, which were not considered in this work; v) investigating how all these strategies interact with varying model architectures and training paradigms remains crucial to fully realize and enhance LLM reasoning capabilities for broader deployment.

\bibliographystyle{unsrtnat}
\bibliography{neurips_2025}


\clearpage

\appendix

\section*{Appendix}

\section{Outline}

This appendix provides supplementary information and additional experimental results to support the main text. The content is organized into four main parts:


        
    
\textbf{Related Work.}
Appendix~\ref{app:related_work} presents a review of existing research configurations for evaluating LLMs' in-context reasoning, discusses key concepts such as In-Context Search, External Search, and Test-Time Scaling, and positions our work within this broader landscape.

\textbf{Prompt Design.} Appendix~\ref{app:setup} presents the specific prompt templates employed for Direct Prompting, Chain of Thought (CoT) Prompting, and Algorithm of Thought (AoT) Prompting, using the Trip Planning task as an illustrative example.
    
\begin{figure}[H] 
    \centering
\hspace{-1mm}\includegraphics[width=100mm]{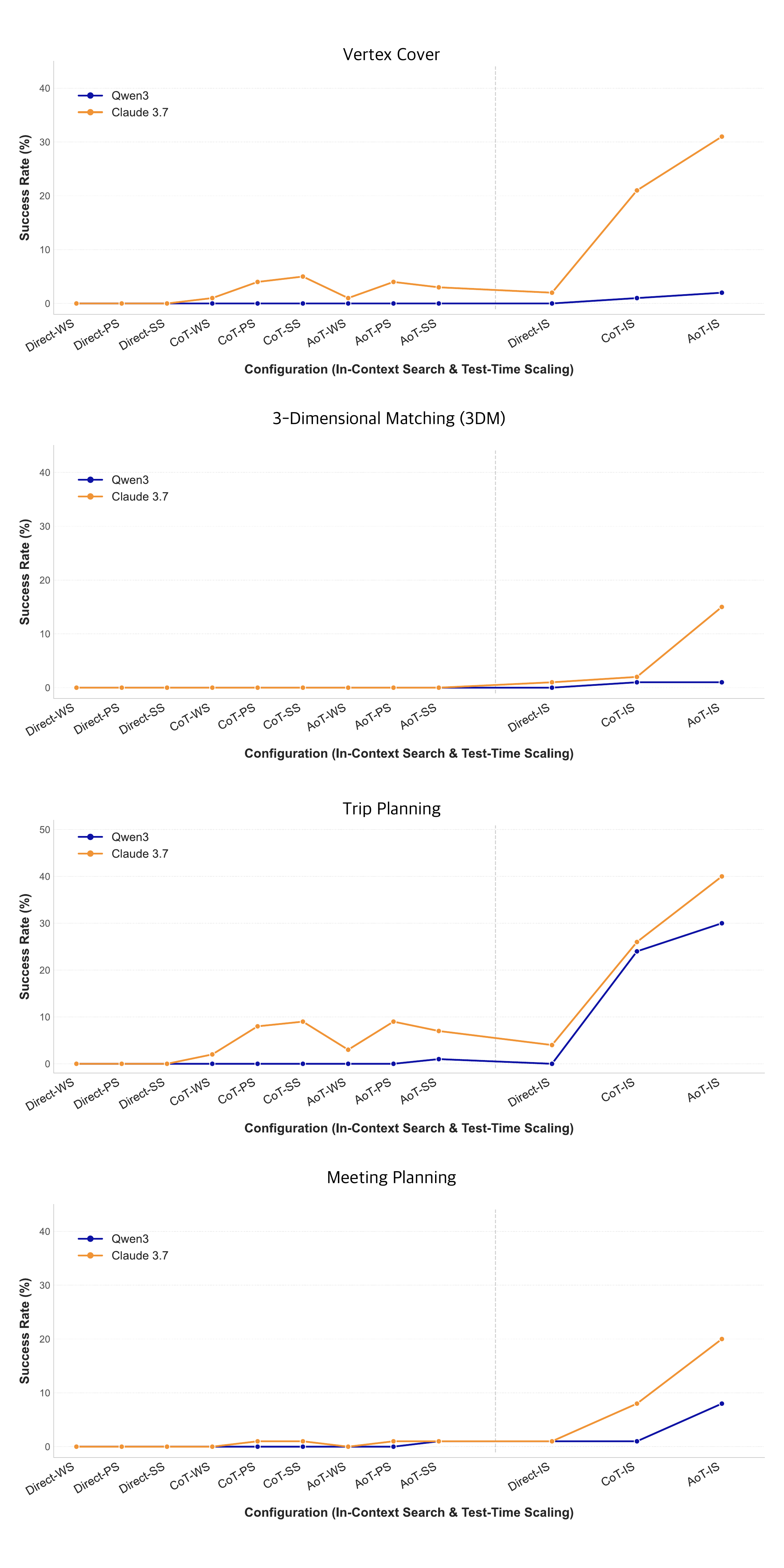}
    \caption{Success rates (\%) of the Qwen3 and Claude 3.7 models across various configurations for four challenging tasks: Vertex Cover, 3-Dimensional Matching (3DM), Trip Planning, and Meeting Planning.}
    \label{fig:line}
    \vspace{-2mm}
\end{figure}

\textbf{Experimental Results.}
Appendix~\ref{app:experiment} presents supplementary experimental results, including detailed tables (e.g., Table~\ref{table:3dm}, Table~\ref{table:meeting_planning}) and line graphs (e.g., Figure~\ref{fig:line}), which expand on the findings discussed in the main text. 
    
\textbf{Theoretical Results.} 
Appendix~\ref{app:theory} illustrates the theoretical foundations of our work, including definitions of Turing Machines, computational complexity classes (such as $\mathsf{P}$, $\mathsf{NP}$, $\mathsf{EXP}$, $\mathsf{NEXP}$, $\mathsf{CoT}(t(n))$, and $\mathsf{AoT}(a(n))$), Transformer model assumptions, and provides formal proofs for key theorems (Theorem~\ref{thm:CoT}, \ref{thm:AoT}, \ref{thm:CoT_exp}, \ref{thm:AoT_exp}, and \ref{thm:core_reasoning_tokens}).




\section{Related Work}\label{app:related_work}

\begin{table}[H]
  \caption{Comparison with existing research configurations evaluating LLMs’ in‑context reasoning; external mechanisms are not considered in this work. Our study is the first to systematically explore the combined potential of Test-Time Scaling and In-Context Search on complex reasoning tasks.} 
  \vspace{3mm}
  \centering
  \begin{adjustbox}{width=0.96\textwidth,center}
    \begin{tabular}{
        >{\raggedright\arraybackslash}p{4cm}@{\hspace{2pt}}
        *{3}{>{\centering\arraybackslash}m{1.3cm}}          
        *{3}{>{\centering\arraybackslash}m{1.8cm}}          
      }
      \toprule
      & \multicolumn{3}{c}{\textbf{Test‑Time Scaling}}
      & \multicolumn{3}{c}{\textbf{In‑Context Search}}\\
      \cmidrule(lr){2-4}\cmidrule(lr){5-7}
      & Parallel Scaling
      & Sequential Scaling
      & \textbf{\textcolor{blue}{Internal Scaling}}
      & \makecell{Direct\\(Zero‑shot/\\Few‑shot/\\Hints)}
      & \makecell{CoT\\(Greedy\\Search)}
      & \hspace*{-8pt}\makecell{\textcolor{blue}{\textbf{AoT}}\\\textcolor{blue}{\textbf{(Algorithmic}}\\\textcolor{blue}{\textbf{Search}})}\\
      \midrule
      \cite{snell2024scaling}   & \cmark & \cmark & \xmark & \cmark & \xmark & \xmark\\
      \cite{maher2025llmpc}     & \cmark & \cmark & \xmark & \cmark & \xmark & \xmark\\
      \cite{iklassov2024self}   & \cmark & \cmark & \xmark & \cmark & \xmark & \xmark\\
      \cite{chen2024can}        & \xmark & \cmark & \xmark & \cmark & \cmark & \xmark\\
      \cite{chen2024unlocking}  & \cmark & \xmark & \cmark & \cmark & \cmark & \xmark\\
      \cite{lee2025evolving}    & \cmark & \cmark & \cmark & \cmark & \xmark & \xmark\\
      \cite{hao2024large}       & \xmark & \cmark & \cmark & \cmark & \cmark & \xmark\\
      \cite{chen2024reprompt}   & \xmark & \cmark & \cmark & \cmark & \xmark & \xmark\\
      \cite{valmeekam2024llms}  & \xmark & \xmark & \cmark & \cmark & \xmark & \xmark\\
      \cite{xie2024revealing}   & \xmark & \xmark & \xmark & \cmark & \xmark & \xmark\\
      \cite{qi2024quantifying}  & \xmark & \xmark & \cmark & \cmark & \xmark & \xmark\\
      \cite{zheng2024natural}   & \xmark & \cmark & \cmark & \cmark & \xmark & \xmark\\
      \cite{wang2025think}      & \cmark & \cmark & \cmark & \cmark & \xmark & \xmark\\
      \cite{sun2025climbing}    & \cmark & \xmark & \cmark & \cmark & \xmark & \xmark\\
      \cite{ge2025innate}       & \xmark & \xmark & \cmark & \cmark & \cmark & \xmark\\
      \cite{hazra2025have}      & \xmark & \xmark & \cmark & \cmark & \xmark & \xmark\\
      \cite{sel2025llms}        & \cmark & \cmark & \xmark & \cmark & \cmark & \cmark\\
      \midrule
      \textbf{Our Work}         & \cmark & \cmark & \cmark & \cmark & \cmark & \cmark\\
      \bottomrule
    \end{tabular}
  \end{adjustbox}
  \label{table:related_work}
\end{table}

\textbf{In-Context Search.} In-context search prompting enables LLMs to learn and execute algorithmic behaviors for searching the solution space of complex problems, often all within a single query. (i) \textbf{Direct Prompting}: This approach involves providing the model with a few problem-solution pairs \citep{wang2020generalizing} without any explicit intermediate reasoning steps in the examples. It relies on the model’s internalized reasoning to bridge from the problem to the solution and generally does not delve into deep task decomposition. Variations include strategies like Automatic Chain-of-Thought (Auto-CoT) prompting~\citep{zhang2022automatic}, and the use of hints before prompting~\citep{fu2024hint}; (ii) \textbf{Chain-of-Thought (CoT) Prompting}: CoT~\citep{wei2022chain} guides LLMs to generate a sequence of intermediate reasoning steps before arriving at a final solution by providing an in-context example that demonstrates a successive, step-by-step solution path (typically representing a \textit{greedy search} without algorithmic operations like backtracking). (iii) \textbf{Algorithm-of-Thought (AoT) Prompting}: AoT~\citep{sel2025llms} is a more advanced in-context prompting strategy that guides LLMs through explicit \textit{algorithmic search} pathways. This is achieved by employing detailed examples demonstrating algorithmic operations. 

\textbf{External Search.} In contrast with in-context search methods, external search involves an external algorithmic pipeline designed for halting, modifying, and then resuming the LLM's generation process, often relying on multiple queries or external tools (e.g. verifiers or state evaluators). Representative works include Tree-of-Thoughts (ToT)~\citep{yao2023tree}, Graph-of-Thoughts (GoT)~\citep{besta2024graph}, and Reasoning-via-Planning (RAP)~\citep{hao2023reasoning} approaches. In this paper, we focus on evaluating and enhancing LLMs' in-context reasoning abilities without the aid of external mechanisms such as additional model training, external reward signals, or explicit process supervision.

\textbf{Test-Time Scaling.} This class of methods aims to boost on-the-fly reasoning capabilities by allocating additional computational resources during inference~\citep{zhang2025and}. Major strategies include (i) \textbf{Parallel Scaling}: improves performance by generating multiple outputs in parallel and then aggregating them, as seen in techniques like Best-of-N (BoN) sampling and Self-Consistency~\citep{wang2022self}; (ii) \textbf{Sequential Scaling}: an iterative method where subsequent computations are explicitly directed based on the results of intermediate steps, allowing the model to refine solutions step-by-step, exemplified by approaches like Self-Refine~\citep{madaan2023self} and ReAct~\citep{yao2023react}; and (iii) \textbf{Internal Scaling}: where a model autonomously determines the computational effort to allocate for reasoning using its internal parameters and a learned policy (typically trained via reinforcement learning~\citep{guo2025deepseek}), enabling more detailed or self-evaluated reasoning chains without external multi-call mechanisms.

\section{Prompt Design} \label{app:setup}

This section shows the different prompting methodologies using Trip Planning task in~\citep{zheng2024natural} as an illustrative example, detailing the specific templates employed for Direct Prompting, CoT Prompting, and AoT Prompting.

\begin{tcolorbox}[
    breakable, 
    colback=white!95!gray,
    colframe=black,
    fonttitle=\bfseries,
    boxrule=0.5mm,
    width=\textwidth,
    before=\par\noindent, 
    after=\par, 
    title={Direct Prompting}
]
    
    \textbf{Problem Description:} 
    \vspace{3mm}
    
    You are an expert at planning trips. You are given a few constraints regarding the cities to visit and the durations of staying at each city. You are also given the flight information between the cities.

    \vspace{3mm}
    \textbf{5-Shot In-Context Examples:}
    \vspace{3mm}
    
    \noindent \#\#\# Task1 \#\#\# \par
    \noindent You plan to visit 6 European cities for 16 days in total. You only take direct flights to commute between cities. You plan to stay in Riga for 3 days. You plan to stay in Edinburgh for 4 days. You want to spend 2 days in Milan. You are going to attend a wedding in Milan between day 4 and day 5. You plan to stay in Copenhagen for 4 days. You plan to visit relatives in Copenhagen between day 5 and day 8. You would like to visit Vilnius for 5 days. You have to attend a workshop in Vilnius between day 10 and day 14. You plan to stay in Brussels for 3 days.\textbackslash{}n\textbackslash{}nHere are the cities that have direct flights:\textbackslash{}nEdinburgh and Copenhagen, Vilnius and Brussels, Copenhagen and Riga, Milan and Vilnius, Milan and Brussels, Edinburgh and Milan, Edinburgh and Riga, Edinburgh and Brussels, Milan and Copenhagen, Copenhagen and Brussels, from Riga to Vilnius, Copenhagen and Vilnius, Milan and Riga, Riga and Brussels.\textbackslash{}n\textbackslash{}nFind a trip plan of visiting the cities for 16 days by taking direct flights to commute between them. \par
    \medskip 

    \noindent \#\#\# Solution1 \#\#\# \par
    \noindent Here is the trip plan for visiting the 6 European cities for 16 days:\textbackslash{}n\textbackslash{}n**Day 1-4:** Arriving in Edinburgh and visit Edinburgh for 4 days.\textbackslash{}n**Day 4:** Fly from Edinburgh to Milan.\textbackslash{}n**Day 4-5:** Visit Milan for 2 days.\textbackslash{}n**Day 5:** Fly from Milan to Copenhagen.\textbackslash{}n**Day 5-8:** Visit Copenhagen for 4 days.\textbackslash{}n**Day 8:** Fly from Copenhagen to Riga.\textbackslash{}n**Day 8-10:** Visit Riga for 3 days.\textbackslash{}n**Day 10:** Fly from Riga to Vilnius.\textbackslash{}n**Day 10-14:** Visit Vilnius for 5 days.\textbackslash{}n**Day 14:** Fly from Vilnius to Brussels.\textbackslash{}n**Day 14-16:** Visit Brussels for 3 days. \par

    \textit{…… (4 more examples )}
    \medskip
    
    \vspace{3mm}
    \textbf{Question:}
    \vspace{3mm}

    \noindent \#\#\# Target Question \#\#\# \par
   You plan to visit 10 European cities for 25 days in total. You only take direct flights to commute between cities. You plan to stay in Berlin for 4 days. You have to attend a workshop in Berlin between day 12 and day 15. You would like to visit Prague for 2 days. You plan to stay in Stuttgart for 5 days. You would like to visit Manchester for 3 days. You want to spend 2 days in Nice. You would like to visit Reykjavik for 2 days. You would like to visit Florence for 3 days. You plan to stay in Vilnius for 5 days. You would like to meet your friends at Vilnius between day 15 and day 19 to tour together. You plan to stay in Oslo for 4 days. You would like to visit Dubrovnik for 4 days. You plan to visit relatives in Dubrovnik between day 1 and day 4.\textbackslash{}n\textbackslash{}nHere are the cities that have direct flights:\textbackslash{}nfrom Reykjavik to Stuttgart, Manchester and Stuttgart, Nice and Berlin, Oslo and Prague, Stuttgart and Berlin, Manchester and Nice, Reykjavik and Oslo, Reykjavik and Prague, Manchester and Prague, Reykjavik and Berlin, Dubrovnik and Manchester, Manchester and Oslo, Manchester and Berlin, Prague and Florence, Berlin and Vilnius, Dubrovnik and Oslo, Nice and Oslo, Berlin and Oslo, Nice and Reykjavik, Vilnius and Oslo.\textbackslash{}n\textbackslash{}nFind a trip plan of visiting the cities for 25 days by taking direct flights to commute between them.

\end{tcolorbox}

\vspace{5mm}

\begin{tcolorbox}[
    breakable, 
    colback=white!95!gray,
    colframe=black,
    fonttitle=\bfseries,
    boxrule=0.5mm,
    width=\textwidth,
    before=\par\noindent, 
    after=\par, 
    title={Chain of Thought Prompting}
]
    
    \textbf{Problem Description:} 
    \vspace{3mm}
    
    You are an expert at planning trips. You are given a few constraints regarding the cities to visit and the durations of staying at each city. You are also given the flight information between the cities.

    \vspace{3mm}
    \textbf{\textcolor{red}{Greedy Search} Thinking Process:}
    \vspace{3mm}
    
\noindent \#\#\# Objective \#\#\# \par
\medskip

\noindent Plan a **16-day** trip that visits the **six** European cities below, using only the direct flights provided. \par
\medskip

\noindent \#\#\# Constraints \#\#\# \par
\medskip

\noindent 1. **Adjacency overlap**\quad The last day of city *i* is also the first day of city *i + 1*. \par
\medskip

\noindent 2. **Stay-length \& fixed-window requirements** \par\smallskip
\noindent City: Riga \\
\noindent Required stay: 3 days \\
\noindent Fixed-day window: -- \par\smallskip

\noindent City: Edinburgh \\
\noindent Required stay: 4 days \\
\noindent Fixed-day window: -- \par\smallskip

\noindent City: Milan \\
\noindent Required stay: 2 days \\
\noindent Fixed-day window: **must cover Day 4--5** (wedding) \par\smallskip

\noindent City: Copenhagen \\
\noindent Required stay: 4 days \\
\noindent Fixed-day window: **must cover Day 5--8** (visit relatives) \par\smallskip

\noindent City: Vilnius \\
\noindent Required stay: 5 days \\
\noindent Fixed-day window: **must cover Day 10--14** (workshop) \par\smallskip

\noindent City: Brussels \\
\noindent Required stay: 3 days \\
\noindent Fixed-day window: -- \par
\medskip

\noindent The sum of **independent days** ($\Sigma$ stay - overlaps) must equal **16** exactly.
\medskip

\noindent 3. **Flights requirements**\quad 

A direct flight exists **only** when explicitly listed: \par\smallskip
\noindent Riga\textrightarrow Vilnius\quad Riga\textrightarrow Edinburgh\quad Riga\textrightarrow Milan\quad Riga\textrightarrow Copenhagen\quad Riga\textrightarrow Brussels \\
\noindent Edinburgh\textrightarrow Copenhagen\quad Edinburgh\textrightarrow Milan\quad Edinburgh\textrightarrow Riga\quad Edinburgh\textrightarrow Brussels \\
\noindent Milan\textrightarrow Copenhagen\quad Milan\textrightarrow Vilnius\quad Milan\textrightarrow Brussels\quad Milan\textrightarrow Riga \\
\noindent Copenhagen\textrightarrow Edinburgh\quad Copenhagen\textrightarrow Riga\quad Copenhagen\textrightarrow Vilnius\quad Copenhagen\textrightarrow Milan\quad Copenhagen\textrightarrow Brussels \\
\noindent Vilnius\textrightarrow Brussels\quad Vilnius\textrightarrow Copenhagen\quad Vilnius\textrightarrow Milan \\
\noindent Brussels\textrightarrow Vilnius\quad Brussels\textrightarrow Copenhagen\quad Brussels\textrightarrow Milan\quad Brussels\textrightarrow Edinburgh\quad Brussels\textrightarrow Riga \par
\medskip

\medskip

\noindent \#\#\# Solution Thinking Process \#\#\# \par
\medskip

\noindent 1.\quad State definition \par\smallskip
\noindent path\quad ordered list of visited cities \\
\noindent UD\quad current independent-day total \\
\noindent start\quad calendar start-day of the last city \\
\noindent used\quad set of visited cities \par
\medskip

\noindent 2.\quad Initialization \par\smallskip
\noindent Initial state: (path=[Edinburgh], UD=4, start=1) \# Edinburgh spans Day 1-4 \par
\medskip

\noindent 3.\quad Greedy Search \par\smallskip
\noindent Step: G0 \\
\noindent Transition tried: **Start Edinburgh** \\
\noindent Calendar preview \& test: Edinburgh Day 1-4 (no window) \\
\noindent Outcome: keep \par\smallskip

\noindent Step: G1 \\
\noindent Transition tried: Edinburgh \textrightarrow{} **Milan** \\
\noindent Calendar preview \& test: Milan Day 4--5 covers wedding (Day 4--5) \\
\noindent Outcome: keep \quad UD = 5 \par\smallskip

\noindent Step: G2 \\
\noindent Transition tried: Milan \textrightarrow{} **Copenhagen** \\
\noindent Calendar preview \& test: Copenhagen Day 5--8 covers relatives (Day 5--8) \\
\noindent Outcome: keep \quad UD = 8 \par\smallskip

\noindent Step: G3 \\
\noindent Transition tried: Copenhagen \textrightarrow{} **Riga** \\
\noindent Calendar preview \& test: Riga Day 8--10 (no hard window broken) \\
\noindent Outcome: keep \quad UD = 10 \par\smallskip

\noindent Step: G4 \\
\noindent Transition tried: Riga \textrightarrow{} **Vilnius** \\
\noindent Calendar preview \& test: Vilnius Day 10--14 covers workshop (Day 10--14) \\
\noindent Outcome: keep \quad UD = 14 \par\smallskip

\noindent Step: G5 \\
\noindent Transition tried: Vilnius \textrightarrow{} **Brussels** \\
\noindent Calendar preview \& test: Brussels Day 14--16 (all constraints now met) \\
\noindent Outcome: **Success** \quad UD = 16 \par
\medskip

\noindent 4.\quad Unique solution path found \par\smallskip
\noindent Edinburgh \textrightarrow{} Milan \textrightarrow{} Copenhagen \textrightarrow{} Riga \textrightarrow{} Vilnius \textrightarrow{} Brussels \par
\medskip

\noindent 5.\quad Output Format \par\smallskip
\noindent Day 1--4 \quad Edinburgh \\
\noindent Day 4--5 \quad Milan \quad \quad \quad \ \ (wedding) \\
\noindent Day 5--8 \quad Copenhagen \ \ (visit relatives) \\
\noindent Day 8--10 \quad Riga \\
\noindent Day 10--14 Vilnius \quad \quad (workshop) \\
\noindent Day 14--16 Brussels \par
\medskip

    \vspace{3mm}
    \textbf{Question:}
    \vspace{3mm}

    \noindent \#\#\# Target Question \#\#\# \par
   You plan to visit 10 European cities for 25 days in total. You only take direct flights to commute between cities. You plan to stay in Berlin for 4 days. You have to attend a workshop in Berlin between day 12 and day 15. You would like to visit Prague for 2 days. You plan to stay in Stuttgart for 5 days. You would like to visit Manchester for 3 days. You want to spend 2 days in Nice. You would like to visit Reykjavik for 2 days. You would like to visit Florence for 3 days. You plan to stay in Vilnius for 5 days. You would like to meet your friends at Vilnius between day 15 and day 19 to tour together. You plan to stay in Oslo for 4 days. You would like to visit Dubrovnik for 4 days. You plan to visit relatives in Dubrovnik between day 1 and day 4.\textbackslash{}n\textbackslash{}nHere are the cities that have direct flights:\textbackslash{}nfrom Reykjavik to Stuttgart, Manchester and Stuttgart, Nice and Berlin, Oslo and Prague, Stuttgart and Berlin, Manchester and Nice, Reykjavik and Oslo, Reykjavik and Prague, Manchester and Prague, Reykjavik and Berlin, Dubrovnik and Manchester, Manchester and Oslo, Manchester and Berlin, Prague and Florence, Berlin and Vilnius, Dubrovnik and Oslo, Nice and Oslo, Berlin and Oslo, Nice and Reykjavik, Vilnius and Oslo.\textbackslash{}n\textbackslash{}nFind a trip plan of visiting the cities for 25 days by taking direct flights to commute between them.

\end{tcolorbox}

\vspace{5mm}

\begin{tcolorbox}[
    breakable, 
    colback=white!95!gray,
    colframe=black,
    fonttitle=\bfseries,
    boxrule=0.5mm,
    width=\textwidth,
    before=\par\noindent, 
    after=\par, 
    title={Algorithm of Thought Prompting}
]
    
    \textbf{Problem Description:} 
    \vspace{3mm}
    
    You are an expert at planning trips. You are given a few constraints regarding the cities to visit and the durations of staying at each city. You are also given the flight information between the cities.

    \vspace{3mm}
    \textbf{\textcolor{red}{Depth-First Search} Thinking Process:}
    \vspace{3mm}
    
\noindent \#\#\# Objective \#\#\# \par
\medskip

\noindent Plan a **16-day** trip that visits the **six** European cities below, using only the direct flights provided. \par
\medskip

\noindent \#\#\# Constraints \#\#\# \par
\medskip

\noindent 1. **Adjacency overlap**\quad The last day of city *i* is also the first day of city *i + 1*. \par
\medskip

\noindent 2. **Stay-length \& fixed-window requirements** \par\smallskip
\noindent City: Riga \\
\noindent Required stay: 3 days \\
\noindent Fixed-day window: -- \par\smallskip

\noindent City: Edinburgh \\
\noindent Required stay: 4 days \\
\noindent Fixed-day window: -- \par\smallskip

\noindent City: Milan \\
\noindent Required stay: 2 days \\
\noindent Fixed-day window: **must cover Day 4--5** (wedding) \par\smallskip

\noindent City: Copenhagen \\
\noindent Required stay: 4 days \\
\noindent Fixed-day window: **must cover Day 5--8** (visit relatives) \par\smallskip

\noindent City: Vilnius \\
\noindent Required stay: 5 days \\
\noindent Fixed-day window: **must cover Day 10--14** (workshop) \par\smallskip

\noindent City: Brussels \\
\noindent Required stay: 3 days \\
\noindent Fixed-day window: -- \par
\medskip

\noindent The sum of **independent days** ($\Sigma$ stay - overlaps) must equal **16** exactly. \medskip

\noindent 3. **Flights requirements**\quad A direct flight exists **only** when explicitly listed: \par\smallskip
\noindent Riga\textrightarrow{}Vilnius\quad Riga\textrightarrow{}Edinburgh\quad Riga\textrightarrow{}Milan\quad Riga\textrightarrow{}Copenhagen\quad Riga\textrightarrow{}Brussels \\
\noindent Edinburgh\textrightarrow{}Copenhagen\quad Edinburgh\textrightarrow{}Milan\quad Edinburgh\textrightarrow{}Riga\quad Edinburgh\textrightarrow{}Brussels \\
\noindent Milan\textrightarrow{}Copenhagen\quad Milan\textrightarrow{}Vilnius\quad Milan\textrightarrow{}Brussels\quad Milan\textrightarrow{}Riga \\
\noindent Copenhagen\textrightarrow{}Edinburgh\quad Copenhagen\textrightarrow{}Riga\quad Copenhagen\textrightarrow{}Vilnius\quad Copenhagen\textrightarrow{}Milan\quad Copenhagen\textrightarrow{}Brussels \\
\noindent Vilnius\textrightarrow{}Brussels\quad Vilnius\textrightarrow{}Copenhagen\quad Vilnius\textrightarrow{}Milan \\
\noindent Brussels\textrightarrow{}Vilnius\quad Brussels\textrightarrow{}Copenhagen\quad Brussels\textrightarrow{}Milan\quad Brussels\textrightarrow{}Edinburgh\quad Brussels\textrightarrow{}Riga \par
\medskip

\noindent \#\#\# Solution Thinking Process \#\#\# \par
\medskip

\noindent 1.\quad State definition \par\smallskip
\noindent path\quad ordered list of visited cities \\
\noindent UD\quad current independent-day total \\
\noindent start\quad calendar start-day of the last city \\
\noindent used\quad set of visited cities \par
\medskip

\noindent 2.\quad Initialization \par\smallskip
\noindent Pick an initial state: (path=[Riga], UD=3, start=1) \# Riga spans Day 1-3 \par
\medskip

\noindent 3.\quad Depth-First Search with pruning \par
\medskip

\noindent 3-A. Riga-rooted subtree (cut in one shot) \par\smallskip 
\noindent Step: A \\
\noindent Transition tried: Riga\textrightarrow{}**Milan** \\
\noindent Calendar preview \& test: Milan Day 3-4 (wedding Day 5 missing) \\
\noindent Outcome: **Prune (window)** \par\smallskip

\noindent Step: B \\
\noindent Transition tried: Riga\textrightarrow{}**Edinburgh** \\
\noindent Calendar preview \& test: Edinburgh Day 3-6 \textrightarrow{} Milan can’t own Day 4-5 \\
\noindent Outcome: **Prune** \par\smallskip

\noindent Step: C \\
\noindent Transition tried: Riga\textrightarrow{}**Brussels** \\
\noindent Calendar preview \& test: Brussels Day 3-5 occupies Day 4-5 \\
\noindent Outcome: **Prune** \par\smallskip

\noindent Step: D \\
\noindent Transition tried: Riga\textrightarrow{}**Copenhagen** \\
\noindent Calendar preview \& test: Copenhagen Day 3-6 (relatives window broken; Milan missing) \\
\noindent Outcome: **Prune** \par\smallskip

\noindent Step: E \\
\noindent Transition tried: Riga\textrightarrow{}**Vilnius** \\
\noindent Calendar preview \& test: Vilnius Day 3-7 (workshop window broken) \\
\noindent Outcome: **Prune** \par 
\medskip 

\noindent No child of the Riga root survives, so the algorithm back-tracks to choose a new start city. \medskip

\noindent 3-B. Edinburgh-rooted search (full trace) \par\smallskip 
\noindent Step: S0 \\
\noindent Transition tried: **Start Edinburgh** \\
\noindent Calendar preview \& test: Edinburgh Day 1-4 \\
\noindent Outcome: keep \par\smallskip

\noindent Step: A \\
\noindent Transition tried: Edinburgh\textrightarrow{}**Copenhagen** \\
\noindent Calendar preview \& test: Copenhagen Day 4-7; relatives window 5-8 not fully covered, wedding lost \\
\noindent Outcome: **Prune (windows)** \par\smallskip

\noindent Step: B \\
\noindent Transition tried: Edinburgh\textrightarrow{}**Brussels** \\
\noindent Calendar preview \& test: Brussels Day 4-6; Milan cannot cover Day 4-5 \\
\noindent Outcome: **Prune (window)** \par\smallskip

\noindent Step: C \\
\noindent Transition tried: Edinburgh\textrightarrow{}**Milan** \\
\noindent Calendar preview \& test: Milan Day 4--5 covers wedding \textrightarrow{} UD = 5 \\
\noindent Outcome: keep \par\smallskip

\noindent Step: C1 \\
\noindent Transition tried: \ldots{}\textrightarrow{}**Brussels** \\
\noindent Calendar preview \& test: Brussels Day 5-7; relatives window lost \\
\noindent Outcome: **Prune** \par\smallskip

\noindent Step: C2 \\
\noindent Transition tried: \ldots{}\textrightarrow{}**Riga** \\
\noindent Calendar preview \& test: Riga Day 5-7; relatives window lost \\
\noindent Outcome: **Prune** \par\smallskip

\noindent Step: C3 \\
\noindent Transition tried: \ldots{}\textrightarrow{}**Vilnius** \\
\noindent Calendar preview \& test: Vilnius Day 5-9; relatives \& workshop windows broken \\
\noindent Outcome: **Prune** \par\smallskip

\noindent Step: C4 \\
\noindent Transition tried: \ldots{}\textrightarrow{}**Copenhagen** \\
\noindent Calendar preview \& test: Copenhagen Day 5--8 covers relatives \textrightarrow{} UD = 8 \\
\noindent Outcome: keep \par\smallskip

\noindent Step: C4a \\
\noindent Transition tried: \ldots{}\textrightarrow{}**Brussels** \\
\noindent Calendar preview \& test: Brussels Day 8-10; workshop window still unmet \\
\noindent Outcome: **Prune** \par\smallskip

\noindent Step: C4b \\
\noindent Transition tried: \ldots{}\textrightarrow{}**Vilnius** \\
\noindent Calendar preview \& test: Vilnius Day 8-12; workshop window 10-14 not fully covered \\
\noindent Outcome: **Prune** \par\smallskip

\noindent Step: C4c \\
\noindent Transition tried: \ldots{}\textrightarrow{}**Riga** \\
\noindent Calendar preview \& test: Riga Day 8--10 \textrightarrow{} UD = 10 \\
\noindent Outcome: keep \par\smallskip

\noindent Step: C4c1 \\
\noindent Transition tried: \ldots{}\textrightarrow{}**Brussels** \\
\noindent Calendar preview \& test: Brussels Day 10-12; workshop window unmet \\
\noindent Outcome: **Prune** \par\smallskip

\noindent Step: C4c2 \\
\noindent Transition tried: \ldots{}\textrightarrow{}**Edinburgh** \\
\noindent Calendar preview \& test: revisit city \\
\noindent Outcome: **Prune (visited)** \par\smallskip

\noindent Step: C4c3 \\
\noindent Transition tried: \ldots{}\textrightarrow{}**Milan** \\
\noindent Calendar preview \& test: revisit city \\
\noindent Outcome: **Prune (visited)** \par\smallskip

\noindent Step: C4c4 \\
\noindent Transition tried: \ldots{}\textrightarrow{}**Vilnius** \\
\noindent Calendar preview \& test: Vilnius Day 10--14 covers workshop \textrightarrow{} UD = 14 \\
\noindent Outcome: keep \par\smallskip

\noindent Step: C4c4a \\
\noindent Transition tried: \ldots{}\textrightarrow{}**Brussels** \\
\noindent Calendar preview \& test: Brussels Day 14--16 \textrightarrow{} UD = 16 (check) \\
\noindent Outcome: **Success** \par
\medskip

\noindent 4.\quad Unique solution path found \par\smallskip
\noindent Edinburgh \textrightarrow{} Milan \textrightarrow{} Copenhagen \textrightarrow{} Riga \textrightarrow{} Vilnius \textrightarrow{} Brussels \par
\medskip

\noindent 5.\quad Output Format \par\smallskip
\noindent Day 1--4 \quad Edinburgh \\
\noindent Day 4--5 \quad Milan \quad \quad \quad \ \ (wedding) \\
\noindent Day 5--8 \quad Copenhagen \ \ (visit relatives) \\
\noindent Day 8--10 \quad Riga \\
\noindent Day 10--14 Vilnius \quad \quad (workshop) \\
\noindent Day 14--16 Brussels \par
\medskip
    
    \vspace{3mm}
    \textbf{Question:}
    \vspace{3mm}

    \noindent \#\#\# Target Question \#\#\# \par
   You plan to visit 10 European cities for 25 days in total. You only take direct flights to commute between cities. You plan to stay in Berlin for 4 days. You have to attend a workshop in Berlin between day 12 and day 15. You would like to visit Prague for 2 days. You plan to stay in Stuttgart for 5 days. You would like to visit Manchester for 3 days. You want to spend 2 days in Nice. You would like to visit Reykjavik for 2 days. You would like to visit Florence for 3 days. You plan to stay in Vilnius for 5 days. You would like to meet your friends at Vilnius between day 15 and day 19 to tour together. You plan to stay in Oslo for 4 days. You would like to visit Dubrovnik for 4 days. You plan to visit relatives in Dubrovnik between day 1 and day 4.\textbackslash{}n\textbackslash{}nHere are the cities that have direct flights:\textbackslash{}nfrom Reykjavik to Stuttgart, Manchester and Stuttgart, Nice and Berlin, Oslo and Prague, Stuttgart and Berlin, Manchester and Nice, Reykjavik and Oslo, Reykjavik and Prague, Manchester and Prague, Reykjavik and Berlin, Dubrovnik and Manchester, Manchester and Oslo, Manchester and Berlin, Prague and Florence, Berlin and Vilnius, Dubrovnik and Oslo, Nice and Oslo, Berlin and Oslo, Nice and Reykjavik, Vilnius and Oslo.\textbackslash{}n\textbackslash{}nFind a trip plan of visiting the cities for 25 days by taking direct flights to commute between them.

\end{tcolorbox}

\section{Experimental Results}
\label{app:experiment}

This section provides additional experimental results for four tasks: Vertex Cover and 3-Dimensional Matching (3DM) from the controlled NP-Hard problems category~\citep{yang2025nondeterministic}, and Trip Planning and Meeting Planning from the complex real-world planning tasks~\citep{zheng2024natural}, supplementing those presented in the main body of the paper and featuring detailed tables and line graphs.

\subsection{Tables}

\begin{table}[H]
\centering
\small 
\caption{Performance on \textit{3-Dimensional Matching (3DM)} (Difficulty Level = 10) for controlled NP-hard task. Qwen3 showed almost no improvement across all methods, possibly because of struggling with complex numerical abstract reasoning.}
\vspace{2mm} 
\label{table:3dm} 

\begin{tabular*}{\textwidth}{@{}p{1.8cm} 
                             @{\extracolsep{\fill}} 
                             >{\RaggedRight\arraybackslash}p{5.0cm} 
                             >{\centering\arraybackslash}p{2.4cm} 
                             @{\extracolsep{\fill}} 
                             >{\centering\arraybackslash}p{2.4cm} 
                             @{\extracolsep{\fill}} 
                             >{\centering\arraybackslash}p{2.6cm}@{}} 
\toprule
\multirow{2}{*}{\textbf{Model}} & \multirow{2}{*}{\textbf{Evaluation Strategy}} & \multicolumn{1}{c}{Direct} & \multicolumn{1}{c}{Greedy Search} & \multicolumn{1}{c}{Depth-First} \\
& & \multicolumn{1}{c}{Prompting} & \multicolumn{1}{c}{(CoT)} & \multicolumn{1}{c}{Search (AoT)} \\
\midrule
\multirow{4}{*}{\textbf{Qwen3}} 
& No Scaling (Base Model) & 0/100 = 0\% & 0/100 = 0\% & 0/100 = 0\% \\
& Parallel Scaling (Best-of-N) & 0/100 = 0\% & 0/100 = 0\% & 0/100 = 0\% \\
& Sequential Scaling (Self-Refine) & 0/100 = 0\% & 0/100 = 0\% & 0/100 = 0\% \\
& Internal Scaling (Thinking Mode) & 0/100 = 0\% & 1/100 = 1\% & 1/100 = 1\% \\
\midrule
\multirow{4}{*}{\textbf{Claude 3.7}} 
& No Scaling (Base Model) & 0/100 = 0\% & 0/100 = 0\% & 0/100 = 0\% \\
& Parallel Scaling (Best-of-N) & 0/100 = 0\% & 0/100 = 0\% & 0/100 = 0\% \\
& Sequential Scaling (Self-Refine) & 0/100 = 0\% & 0/100 = 0\% & 0/100 = 0\% \\
& Internal Scaling (Thinking Mode) & 1/100 = 1\% & 2/100 = 2\% & 15/100 = \textbf{15\%} \\
\bottomrule
\end{tabular*}
\end{table}

\begin{table}[htbp] 
\centering
\small 
\caption{Performance on \textit{Meeting Planning} (Difficulty Level = 10) for complex real-world planning. When the complexity of numerical input in language-based planning is disentangled, Qwen3 improved.}
\vspace{2mm} 
\label{table:meeting_planning}

\begin{tabular*}{\textwidth}{@{}p{1.8cm} 
                             @{\extracolsep{\fill}} 
                             >{\RaggedRight\arraybackslash}p{5.0cm} 
                             >{\centering\arraybackslash}p{2.4cm} 
                             @{\extracolsep{\fill}} 
                             >{\centering\arraybackslash}p{2.4cm} 
                             @{\extracolsep{\fill}} 
                             >{\centering\arraybackslash}p{2.6cm}@{}} 
\toprule
\multirow{2}{*}{\textbf{Model}} & \multirow{2}{*}{\textbf{Evaluation Strategy}} & \multicolumn{1}{c}{Direct} & \multicolumn{1}{c}{Greedy Search} & \multicolumn{1}{c}{Depth-First} \\
& & \multicolumn{1}{c}{Prompting} & \multicolumn{1}{c}{(CoT)} & \multicolumn{1}{c}{Search (AoT)} \\
\midrule
\multirow{4}{*}{\textbf{Qwen3}} 
& No Scaling (Base Model) & 0/100 = 0\% & 0/100 = 0\% & 0/100 = 0\% \\
& Parallel Scaling (Best-of-N) & 0/100 = 0\% & 0/100 = 0\% & 0/100 = 0\% \\
& Sequential Scaling (Self-Refine) & 0/100 = 0\% & 0/100 = 0\% & 1/100 = 1\% \\
& Internal Scaling (Thinking Mode) & 0/100 = 0\% & 1/100 = 1\% & 8/100 = \textbf{8\%} \\
\midrule
\multirow{4}{*}{\textbf{Claude 3.7}} 
& No Scaling (Base Model) & 0/100 = 0\% & 0/100 = 0\% & 0/100 = 0\% \\
& Parallel Scaling (Best-of-N) & 0/100 = 0\% & 1/100 = 1\% & 1/100 = 1\% \\
& Sequential Scaling (Self-Refine) & 0/100 = 0\% & 1/100 = 1\% & 1/100 = 1\% \\
& Internal Scaling (Thinking Mode) & 1/100 = 1\% & 8/100 = 8\% & 20/100 = \textbf{20\%} \\
\bottomrule
\end{tabular*}
\end{table}
\FloatBarrier

\subsection{Line Graphs}

\section{Theoretical Results}
\label{app:theory}

\subsection{Preliminaries}
In this section, we provide theoretical foundations for turing machine, computational complexity classes, and decoder-only transformers.

\begin{definition}[Turing Machine]
A (general) Turing machine \(M\) is defined as a tuple
\[
  M = (Q, \Sigma, \Gamma, \delta, q_0, q_{accept}, q_{reject}),
\]
where
\begin{itemize}[leftmargin=4mm]
  \item \(Q\) is a finite set of states.
  \item \(\Sigma\) is a finite input alphabet, not containing the blank symbol \(\sqcup\).
  \item \(\Gamma\) is a finite tape alphabet with \(\Sigma \subset \Gamma\) and \(\sqcup \in \Gamma\).
  \item \(q_0 \in Q\) is the start state; \(q_{accept},q_{reject}\in Q\) are the unique halting states, \(q_{accept}\neq q_{reject}\).
  \item \(\delta\) is the transition function, which differs for deterministic vs.\ nondeterministic machines:
    \[
      \begin{aligned}
      &\text{if }M\text{ is deterministic:}
      &&\delta: Q\times\Gamma \;\to\; Q\times\Gamma\times\{L,R\}, \\
      &\text{if }M\text{ is nondeterministic:}
      &&\delta: Q\times\Gamma \;\to\; \mathcal{P}\bigl(Q\times\Gamma\times\{L,R\}\bigr).
      \end{aligned}
    \]
     $L$ signifies a left movement and $R$ a right movement of the tape head. In the deterministic case, exactly one action is specified for each \((q,a)\).  In the nondeterministic case, \(\delta(q,a)\) is a finite set of possible moves.
\end{itemize}
On input \(w\in\Sigma^*\), the machine initially places \(w\) on the tape, with the head at the leftmost cell and state \(q_0\).  It then follows transitions until it reaches either \(q_{accept}\) (accept) or \(q_{reject}\) (reject).

\smallskip

\begin{itemize}[leftmargin=4mm]
  \item A Deterministic Turing Machine (DTM) \(M\) \emph{decides} a language \(L\subseteq\Sigma^*\) if on every input \(w\), it halts in either \(q_{accept}\) or \(q_{reject}\), and accepts exactly those \(w\in L\).
  \item An Non-deterministic Turing Machine (NTM) \(M\) \emph{decides} \(L\) if on every \(w\in L\) there is at least one accepting branch (reaching \(q_{accept}\)), and on every \(w\notin L\) no branch accepts.
  \item The \emph{time complexity} of \(M\), denoted \(T_M(n)\), is the maximum number of steps (over all computation paths, in the nondeterministic case) on inputs of length \(n\).

\end{itemize}

\end{definition}

\vspace{2mm}

\begin{definition}[Deterministic and Nondeterministic Time Classes]
\label{def:det_class}
Let $t:\mathbb{N}\to\mathbb{N}$ be a time‐constructible function.
\[
  \mathsf{TIME}(t(n))
    = \{\,L \mid \exists\ \text{DTM }M\ \text{that decides }L
      \text{ in }O(t(n))\text{ time}\}.
\]

A language $L$ is in $\mathsf{TIME}(t(n))$ precisely when there is a deterministic Turing machine that, on every input $w$ of length $n$, halts within $O(t(n))$ steps and correctly decides whether $w\in L$.  In other words, $\mathsf{TIME}(t(n))$ captures exactly those problems solvable by a DTM within the specified time bound.
\[
  \mathsf{NTIME}(t(n))
    = \{\,L \mid \exists\ \text{NTM }M\ \text{that decides }L
      \text{ in }O(t(n))\text{ time}\}.
\]

Equivalently, $L\in\mathsf{NTIME}(t(n))$ if there is a nondeterministic Turing machine which, on each branch and for each input of length $n$, halts within $O(t(n))$ steps and accepts exactly the strings in $L$.  One can also view this via a polynomial‐time verifier: there is a deterministic machine $V$ and for every $w\in L$ a “witness” $u$ of length $O(t(n))$ such that $V(\langle w,u\rangle)$ accepts in $O(t(n))$ time.

\end{definition}

\vspace{2mm}

\begin{definition}[Polynomial‐Time and Exponential‐Time Classes]
\label{def:time_class}
Using the above classes, we define the most common time‐based complexity classes:
\[
  \mathsf{P}
    = \bigcup_{k\in\mathbb{N}}\mathsf{TIME}(n^k).
\]
A language $L$ is in $\mathsf{P}$ if and only if it can be decided by some deterministic Turing machine in time polynomial in the input length.  Equivalently, there exists $k$ such that on every input of length $n$, the machine halts in at most $O(n^k)$ steps.
\[
  \mathsf{NP}
    = \bigcup_{k\in\mathbb{N}}\mathsf{NTIME}(n^k).
\]
A language $L$ is in $\mathsf{NP}$ if there is a nondeterministic machine running in polynomial time that accepts exactly the strings in $L$.  Equivalently, membership in $L$ can be certified by a polynomial‐length witness which a polynomial‐time deterministic verifier checks.
\[
  \mathsf{EXP}
    = \bigcup_{k\in\mathbb{N}}\mathsf{TIME}\bigl(2^{n^k}\bigr).
\]
The class $\mathsf{EXP}$ contains those problems solvable by a deterministic Turing machine in time bounded by some exponential function $2^{n^k}$.  It strictly generalizes $\mathsf{P}$ by allowing much larger time bounds.
\[
  \mathsf{NEXP}
    = \bigcup_{k\in\mathbb{N}}\mathsf{NTIME}\bigl(2^{n^k}\bigr).
\]
Similarly, $\mathsf{NEXP}$ consists of languages decidable by a nondeterministic Turing machine in exponential time, or equivalently those with exponentially long witnesses checkable in exponential time.
\end{definition}

\vspace{2mm}

\begin{definition}[Polynomial-Time Reducibility ($\le_P$)]
A language $L_1$ is said to be polynomial-time reducible to a language $L_2$, denoted $L_1 \le_P L_2$, if there exists a function $f: \Sigma^*  \rightarrow \Sigma^* $, computable by a DTM in polynomial time, such that for every $w \in \Sigma^* $:
$$ w \in L_1 \iff f(w) \in L_2. $$
\end{definition}

\vspace{2mm}

\begin{definition}[The Complexity Class $\mathsf{NP\text{-}Hard}$]
A problem (or language) $H$ is classified as $\mathsf{NP\text{-}Hard}$ if, for every language $L' \in \mathsf{NP}$, $L' \le_P H$. An $\mathsf{NP\text{-}Hard}$ problem is at least as difficult as any problem in $\mathsf{NP}$. If an $\mathsf{NP\text{-}Hard}$ problem $H$ is also a member of $\mathsf{NP}$, then $H$ is termed $\mathsf{NP\text{-}Complete}$.
\end{definition}

\vspace{2mm}

\begin{definition}[The Complexity Class $\mathsf{CoT}\left(t\left(n\right)\right)$]
\label{def:cot-complexity}

Following~\citep{merrill2023expressive}, we denote by $\mathsf{CoT}\left(t\left(n\right)\right)$ the set of languages $L$, such that there is some transformer $f$ that recognizes $L$ in $t\left(n\right)$ decoding steps.
\end{definition}

\vspace{2mm}

\begin{definition}[The Complexity Class $\mathsf{AoT}\left(a\left(n\right)\right)$]
\label{def:aot-complexity}
Following~\citep{sel2023algorithm}, we denote by $\mathsf{AoT}\left(a\left(n\right)\right)$, the set of languages $L$, such that there is some transformer $f$ and $b>1$, so that $f$ that recognizes $L$ along some successful computational trace of $a\left(n\right)$ decoding steps, using $\mathcal{O}\left(b^{a\left(n\right)}\right)$ intermediate tokens to generate a tree of guesses (e.g. when using DFS).
\end{definition}

\vspace{2mm}

\begin{remark}[On the Total Computational Effort for AoT]
\label{rem:aot-exp-exploration}
It is important to distinguish between the length of the successful computational trace as defined in $\mathsf{AoT}(a(n))$ and the total number of intermediate tokens that an AoT-prompted Transformer might generate when engaged in an exhaustive search strategy to guarantee finding such a trace for a challenging problem instance (e.g., an $\mathsf{NP}$-hard problem).

\begin{itemize}[leftmargin=4mm, itemsep=1mm]
    \item For a language $L \in \mathsf{NP}$, while Theorem~\ref{thm:AoT} states that an accepting AoT trace will have $a(n) = \mathtt{poly}(n)$ tokens, a Transformer prompted to exhaustively search for this trace (e.g., by systematically exploring all possible certificates up to a certain polynomial length) might generate a total number of intermediate tokens that is exponential in $n$, potentially $\mathcal{O}(\exp(\mathtt{poly}(n)))$, before the correct polynomial-length accepting trace is found or all possibilities are exhausted. This reflects the inherent difficulty of search in $\mathsf{NP}$ problems under the exponential time hypothesis~\citep{Impagliazzo1999exptimehypothesis}. The $\mathsf{AoT}(\mathtt{poly}(n))$ classification is based on the length of the specific path, not the effort to find it.

    \item For $\mathsf{AoT}(\exp(n))$, where an accepting trace itself could be exponentially long (e.g., for problems in $\mathsf{NEXP}$ where certificates can be exponentially long), a Transformer prompted to guarantee finding such an exponentially long solution trace through an exhaustive search might generate a total number of intermediate tokens that is doubly exponential in $n$, potentially on the order of $\mathcal{O}(\exp(\exp(\mathtt{poly}(n))))$. This is because the search space could be exponential in the length of the already exponentially long certificate.
\end{itemize}

While generating token sequences representing such exhaustive searches would clearly be computationally infeasible, the empirical results presented in~\ref{sec:exp} suggest that in practice, LLMs tend to produce good guesses sufficiently early in their search to find solutions before exhausting compute limitations in many instances. 

\end{remark}

\vspace{2mm}

\begin{assumption} [$\mathsf{P} \neq \mathsf{NP}$, $\mathsf{EXP} \neq \mathsf{NEXP}$]
Throughout this paper, we adopt the standard and widely accepted conjecture that $\mathsf{P} \neq \mathsf{NP}$ and $\mathsf{EXP} \neq \mathsf{NEXP}$.
\end{assumption}

\vspace{1mm}

\begin{assumption} [Decoder-Only Transformers]
\label{def:transformer}
We consider transformers with decoder-only architecture operating under several key assumptions according to ~\citep{chen2024theoretical, merrill2023expressive}: 

\begin{itemize}[leftmargin=4mm]
    \item \textbf{Constant Depth:} The number of Transformer layers, $L_{depth}$, is a constant, independent of the input length $n$.
    $$ L_{depth} = \mathcal{O}(1) $$
    This implies that the intrinsic sequential processing capability of the base Transformer (without CoT) does not scale with $n$.

    \item \textbf{Logarithmic Precision for Parameters and Activations:} The model operates with finite numerical precision for its weights $W$ and activations $A$. This precision, $P_{bits}$, is logarithmic with respect to the combined length of the input $n$ and the Chain of Thought $t(n)$.
    $$ P_{bits}(W, A) = \mathcal{O}(\mathtt{log}(n+t(n))) $$
    This constraint is important for simulating Transformers on Turing machines efficiently.

    \item \textbf{Logarithmic Embedding Dimension:} The dimensionality of token embeddings, $d_{emb}$, is logarithmic with respect to the input length $n$.
    $$ d_{emb} = \mathcal{O}(\mathtt{log} n) $$
    This keeps the model size relatively small in terms of its width.

    \item \textbf{Strict Causal Masking:} The self-attention mechanism ensures that the computation for a token at sequence position $i$ can only attend to tokens at positions $j$ such that $j < i$. If $M_{mask}$ is the attention mask, then for attention scores $\alpha_{i,j}$:
    $$ \alpha_{i,j} \neq 0 \implies j < i $$
    This is inherent to autoregressive generation in decoder-only models.

    \item \textbf{Advanced Normalization Schemes:} The architecture is assumed to employ specific layer normalization variants, such as projected pre-norm (where a learnable linear projection $P_{proj}$ is applied to the input $h_{sub}$ of a sublayer before layer normalization $\text{LN}$) or multi-pre-norm.
    $$ \mathrm{Output}_{sub} = \mathrm{Sublayer}(\text{LN}(P_{proj}(h_{sub}))) \quad (\text{Illustrative for projected pre-norm}) $$
    These are noted as important for enabling certain constructive proofs.

\item \textbf{Saturated Attention:}  
For every attention head, the score matrix $\alpha\in[0,1]^{n\times n}$ satisfies  
\[
\alpha_{i,j}\in\Bigl\{\,0,\;\frac{1}{k}\Bigr\}\quad\text{for some }k\in\{1,\dots,n\}, 
\qquad
\sum_{j=1}^{n}\alpha_{i,j}=1\quad\forall i.
\] 
Thus each token $i$ either ignores a position ($\alpha_{i,j}=0$) or attends with equal weight $1/k$ to exactly $k$ positions. Uniform attention ($k=n$) and hard attention ($k=1$) are special cases. This “averaging hard attention’’ regime yields deterministic routing that is convenient for expressivity analyses.

\item \textbf{Projected Pre‑Norm:}  
Let $v\in\mathbb{R}^m$ be the sub‑layer input and $M\in\mathbb{R}^{m\times m}$ a learnable projection.  
\[
\mathrm{proj\_layer\_norm}(v;M)\;:=\;\mathrm{layer\_norm}\bigl(Mv\bigr).
\]
By projecting first and normalizing afterward, a layer can isolate and renormalize any linear subspace of its hidden state, effectively simulating multiple independent pre‑norm paths with modest extra depth.

\item \textbf{Layer‑Norm Hash:}  
For scalars $x,y\in\mathbb{R}$ define  
\[
\phi(x,y)\;:=\;\mathrm{layer\_norm}\left([\,x,\;y,\;-x,\;-y\,]\right)\in\mathbb{R}^{4}.
\]
Key properties used throughout the lower‑bound constructions:
\begin{enumerate}[label=(\alph*), itemsep=0.5mm]
  \item Scale‑invariance:\quad $\phi \bigl(x/i,\,1/i\bigr)=\phi(x,1)$ for any $i>0$.
  \item Exact‑match test:\quad $\phi(x,1)\cdot\phi(k,1)=1\;\Longleftrightarrow\;x=k$.
\end{enumerate}
These allow attention to perform reliable equality checks and value retrieval across positions, even when earlier averaging operations have rescaled the stored pairs.

\end{itemize}

\end{assumption}

\vspace{2mm}
\subsection{Proof of Theorem~\ref{thm:CoT}: \emph{$\mathsf{CoT}(\mathtt{poly}(n)) = \mathsf{P}$}~\cite{merrill2023expressive}}\label{subsec:proof-CoT}
\begin{proof}

The proof is demonstrated in two parts:

\textbf{Part I: $\mathsf{P} \subseteq \mathsf{CoT}(\mathtt{poly}(n))$}

This inclusion is established by demonstrating that a CoT-augmented Transformer, under specific architectural assumptions, can simulate any DTM operating within a polynomial time bound.

\textbf{Theorem 2 in~\cite{merrill2023expressive}:} Let $M$ be a DTM that, on input length $1 + n$, runs for at most $t(n)$ steps (at most polynomial). There is a decoder-only projected pre-norm transformer with strict causal saturated attention (with or without positional encodings, as stated in saturated attention assumption in Assumption~\ref{def:transformer}) that, on input $x$, takes $t(n)$ decoding steps and then, with $|M(x)|$ additional steps, outputs $M(x)$.

\textbf{Corresponding Proof.} The proof constructs a Transformer that simulates the DTM $M$ by dedicating each decoding step to a single computational step of $M$. The Transformer’s sequence of generated CoT tokens serves to encode the evolving configuration of $M$. This configuration includes $M$’s current state, the symbols on its tapes, and the positions of its tape heads. Specifically, at each step $j$ of the simulation (corresponding to the $j$-th CoT token), the Transformer may store a ``diff'' representing the changes to $M$’s tape and state from step $j - 1$ to $j$.

A crucial component of this simulation is the ``layer-norm hash'' mechanism. This mechanism allows the Transformer, at any step $i$, to retrieve information corresponding to specific tape cells or past DTM states from the previously generated CoT sequence (tokens $1, \dots, i - 1$). This is achieved under the assumption of strict causal masking in Assumption~\ref{def:transformer}, effectively simulating random access to $M$'s tape. For instance, to determine the symbol $\gamma_i^\tau$ on tape $\tau$ at head position $h_i^\tau$ during the $i$-th CoT step, the Transformer attends to the CoT history. The layer-norm hash $\phi(h_i^\tau / i, 1 / i) = \phi(h_i^\tau, 1)$ is used. An attention query, such as $\langle \phi(h_i^\tau, 1), e_1 \rangle$, can be matched against keys derived from previous CoT steps, e.g., $\langle \phi(h_j^\tau, 1), -\phi(f(j), 1) \rangle$ (where $j < i$), to identify and retrieve the most recent symbol written to cell $h_i^\tau$.

The step-by-step simulation proceeds as follows: For each computational step of $M$, the Transformer performs one CoT generation step (say, step $i$, for $i > n$, where $n$ is the input length). First, the Transformer reconstructs the DTM's current configuration. The DTM state $q_{i-n-1}$ is derived from the $(i-1)$-th CoT token. The symbols under $M$'s tape heads are retrieved from the CoT history using the layer-norm hash mechanism. For the input tape, the symbol $\sigma_{h_i^0}$ at head position $h_i^0$ is retrieved via attention, potentially using a query like $\langle \phi(h_i^0 / i, -1), 1 \rangle$. For work and output tapes, the symbol $\gamma_i^\tau$ at head position $h_i^\tau$ is determined by finding the latest ``diff'' token $\delta_{j-n-1}$ (for some $j < i$) in the CoT history that corresponds to a write operation at cell $h_j^\tau$.

Second, the Transformer’s feed-forward networks implement $M$’s transition function $\delta$. Given the reconstructed state $q_{i-n-1}$ and the symbols read from the tapes $(\sigma_{h_i^0}, \gamma_i^1, \dots, \gamma_i^k)$, these networks compute the new state of $M$, the symbols to be written to the tapes, and the movements of the tape heads.

Finally, this output (new state, symbols written, head movements) is encoded and emitted as the next CoT token, $\delta_{i-n}$. This cycle is repeated for $t(n)$ steps, thereby simulating $t(n)$ steps of $M$. After $t(n)$ simulation steps, if $M$ has halted, the Transformer can then output $M(x)$ using $|M(x)|$ additional decoding steps by retrieving the contents of $M$'s output tape from the CoT sequence.

The proof of Theorem 2 above, demonstrating the capability of a Transformer to simulate a DTM step-for-step, directly leads to the following corollary relating their computational power:

\textbf{Corollary 2.1 in~\cite{merrill2023expressive}:}
\( \mathsf{TIME}(t(n)) \subseteq \mathsf{CoT}(t(n)). \)

With Theorem 2 and Corollary 2.1 established, we can now prove the first part of Theorem~\ref{thm:CoT}. Let $L$ be an arbitrary language in the class $\mathsf{P}$. By the definition of $\mathsf{P}$, there exists a DTM $M_L$ and a polynomial function $p_L(k)$ such that $M_L$ decides $L$, and for any input $w$ of length $n = |w|$, $M_L$ halts in $O(p_L(n))$ time. This implies $L \in \mathsf{TIME}(p_L(n))$.
To show that $L \in \mathsf{CoT}(\mathtt{poly}(n))$, we utilize a CoT-augmented Transformer $M_{CoT}$ configured to simulate $M_L$ as per the construction in Theorem 2. We set the number of intermediate CoT steps generated by $M_{CoT}$ to be $t(n) = p_L(n)$. Since $p_L(n)$ is a polynomial function of $n$, it follows that $t(n) = \mathtt{poly}(n)$.
According to Corollary 2.1, if a language is in $\mathsf{TIME}(t(n))$, then it is also in $\mathsf{CoT}(t(n))$. Since $L \in \mathsf{TIME}(p_L(n))$, by applying Corollary 2.1 with $t(n) = p_L(n)$, we have $L \in \mathsf{CoT}(p_L(n))$. Given our choice $t(n) = p_L(n) = \mathtt{poly}(n)$, it follows that $L \in \mathsf{CoT}(\mathtt{poly}(n))$.
As $L$ was an arbitrary language in $\mathsf{P}$, this establishes the inclusion $\mathsf{P} \subseteq \mathsf{CoT}(\mathtt{poly}(n))$.

\textbf{Part II: $\mathsf{CoT}(\mathtt{poly}(n)) \subseteq \mathsf{P}$}

This inclusion is demonstrated by showing that the computation performed by a CoT-augmented Transformer can be simulated by a DTM in polynomial time.

\textbf{Theorem 3 in~\cite{merrill2023expressive}:} The relationship between CoT and Turing machine time complexity is given by:
\( \mathsf{CoT}(t(n)) \subseteq \widetilde{\mathsf{TIME}}(n^2 + t(n)^2). \)
(The tilde in $\widetilde{\mathsf{TIME}}$ hides polylogarithmic factors, specifically $\mathtt{log}^k(n+t(n))$ for some constant $k$).

\textbf{Corresponding Proof.} The proof proceeds by constructing a DTM that simulates the operations of a CoT-augmented Transformer. The DTM simulates the Transformer for each of its $t(n)$ intermediate CoT generation steps, plus one final step to produce the answer.

To generate a single CoT token, the Transformer performs a forward pass. This pass involves several stages: first, the current input sequence, which consists of the original input of length $n$ and all previously generated CoT tokens up to that point, is embedded. Let $k$ be the number of CoT tokens generated so far; the length of this sequence is $N_k = n+k$. This embedded sequence is then processed through a constant number of layers (see constant depth assumption in Assumption~\ref{def:transformer}). Each layer typically comprises a self-attention mechanism followed by feed-forward networks. Finally, a classification layer selects the next token to be generated.

The DTM simulates each of these operations. A key component in the complexity analysis is the self-attention mechanism. For a sequence of length $N_k$, each self-attention layer performs computations that are roughly quadratic in $N_k$, i.e., $O(N_k^2)$, multiplied by factors related to the model's embedding dimension and number of attention heads (which are assumed to be constant or at most logarithmic in $N_k$ according to Assumption~\ref{def:transformer}). The architectural assumptions include log-precision arithmetic, meaning that numbers are represented with $O(\mathtt{log}(n+t(n)))$ bits. Consequently, each arithmetic operation (such as addition or multiplication) on these numbers can be simulated by the DTM in $O(\mathtt{polylog}(N_k))$ time.

The DTM performs a total of approximately $n+t(n)$ such forward passes (one for each token processed or generated). The $j$-th forward pass in this sequence processes an input of length $j$. The cost of simulating the $j$-th forward pass is polynomial in $j$ and polylogarithmic in the precision. Summing the costs for all forward passes, where the $j$-th pass iterates over $j$ key-value pairs for attention, leads to the overall simulation time. The total time for the DTM to simulate all $n+t(n)$ forward passes is bounded by $\tilde{O}(n^2 + t(n)^2)$. This complexity arises from summing the costs of individual forward passes, roughly expressed as the sum $\sum_{j=1}^{n+t(n)} O(j \cdot (\text{complexity of processing one item including attention lookups}))$, where the per-item processing complexity considering attention over $j$ items is itself $O(j \cdot \mathtt{polylog}(j))$. This sum is $O((n+t(n))^2 \cdot \mathtt{polylog}(n+t(n)))$, which is denoted as $\tilde{O}(n^2+t(n)^2)$.

With Theorem 3 established, we can now prove the second part of Theorem~\ref{thm:CoT}. Let $L$ be an arbitrary language in $\mathsf{CoT}(\mathtt{poly}(n))$. By definition, this implies that there exists a CoT-augmented Transformer, $M_{CoT}$, and a polynomial function $p_M(n)$ such that $M_{CoT}$ decides the language $L$ by generating $t(n) = p_M(n)$ intermediate CoT steps.
According to Theorem 3, the Transformer $M_{CoT}$, which generates $t(n)$ CoT tokens for an input of length $n$, can be simulated by a deterministic Turing machine in $\tilde{O}(n^2 + (t(n))^2)$ time. We substitute $t(n) = p_M(n)$ into this time bound. The simulation time for $M_{CoT}$ by the DTM thus becomes $\tilde{O}(n^2 + (p_M(n))^2)$.
Since $p_M(n)$ is a polynomial function of $n$, its square, $(p_M(n))^2$, is also a polynomial in $n$. Consequently, the expression $n^2 + (p_M(n))^2$ represents a polynomial function of $n$. The $\tilde{O}$ notation encapsulates polylogarithmic factors, which do not alter the overall polynomial nature of this time bound.
Therefore, the DTM simulates $M_{CoT}$ in time that is polynomial with respect to the input length $n$. This signifies that the language $L$, decided by $M_{CoT}$, belongs to the complexity class $\mathsf{P}$.
As $L$ was an arbitrary language in $\mathsf{CoT}(\mathtt{poly}(n))$, we conclude that $\mathsf{CoT}(\mathtt{poly}(n)) \subseteq \mathsf{P}$.

Combining inclusions from \textbf{Part I} and \textbf{Part II}, it directly follows that:
\( \mathsf{CoT}(\mathtt{poly}(n)) = \mathsf{P}. \)

\end{proof}

\subsection{Proof of Theorem~\ref{thm:AoT}: \emph{$\mathsf{AoT}(\mathtt{poly}(n)) = \mathsf{NP}$}}\label{subsec:proof-AoT}
\begin{proof}

The proof is demonstrated in two parts: $\mathsf{NP} \subseteq \mathsf{AoT}(\mathtt{poly}(n))$ and $\mathsf{AoT}(\mathtt{poly}(n)) \subseteq \mathsf{NP}$. We rely on the standard definitions of these complexity classes and the specified architectural assumptions for the Transformer (Assumption~\ref{def:transformer}).

\textbf{Part I: $\mathsf{NP} \subseteq \mathsf{AoT}(\mathtt{poly}(n))$}

This inclusion is established by demonstrating that an AoT-prompted Transformer can decide any language $L \in \mathsf{NP}$. Let $L$ be an arbitrary language in the complexity class $\mathsf{NP}$. By the verifier-based definition of $\mathsf{NP}$, there exists a polynomial function $p_c(k)$ (bounding the certificate length) and a DTM $V$ (the verifier), along with another polynomial function $p_V(k)$ (bounding the verifier's runtime), such that for any input string $w$ of length $n = |w|$:
\[
w \in L \iff \exists u \in \Sigma^* \text{ such that } |u| \le p_c(n) \text{ and } V \text{ accepts the pair } \langle w, u \rangle \text{ within } p_V(|w|+|u|) \text{ time}.
\]
The string $u$ is the certificate (or witness) for the membership of $w$ in $L$. Since $|u| \le p_c(n)$, the length of the input to $V$, $|w|+|u|$, is bounded by $n+p_c(n)$. Consequently, the runtime of $V$, $p_V(n+p_c(n))$, is also a polynomial in $n$. Let this polynomial runtime of $V$ be denoted as $p'_V(n)$.

We will construct an AoT-prompted Transformer, $M_{AoT}$, that decides $L$. The AoT process for $M_{AoT}$ will simulate the nondeterministic guess of a certificate $u$ through guided exploration, and then deterministically verify this certificate using a simulation of the DTM $V$.

\textit{1. Certificate Generation (Simulating Nondeterministic Guess via AoT Exploration):}
AoT prompting enables the Transformer to explore various computational pathways. This capability is harnessed to search for or generate a candidate certificate $u$. The in-context examples provided to $M_{AoT}$ are designed to demonstrate an effective search strategy for such certificates (e.g., by showing how to construct them incrementally, explore a constrained search space, or emulate steps of a search algorithm).

If $w \in L$, a valid certificate $u$ with $|u| \le p_c(n)$ exists and, by Definition~\ref{def:aot-complexity}, the search space is sufficiently large to include all such candidates, so in particular $u$. The AoT-prompted Transformer, through its explorative generation capabilities, can produce a sequence of intermediate thought tokens representing such a candidate certificate $u$. These tokens are generated autoregressively. The segment of an AoT computational path corresponding to the generation of $u$ would consist of a number of tokens proportional to $|u|$. Assuming each symbol of $u$ can be represented by a constant number of tokens (or a number of tokens polylogarithmic in $n$, fitting within the model's tokenization scheme), this requires at most $k_u = \mathcal{O}(p_c(n))$ tokens. This phase effectively emulates the guessing or existential quantification aspect of $\mathsf{NP}$. The AoT framework allows the model to try different potential certificates if earlier attempts do not lead to verification, using backtracking and exploration as guided by its prompting.

\textit{2. Verification (Simulating Verifier DTM $V$):}
Once a candidate certificate $u$ (represented by at most $k_u$ tokens) has been generated as part of an AoT computational path, the Transformer then simulates the execution of the verifier DTM $V$ on the input pair $\langle w, u \rangle$.
The verifier $V$ runs in $p'_V(n)$ time on this input. According to Theorem 2 in~\cite{merrill2023expressive} (as cited in the proof of Theorem~\ref{thm:CoT}), a decoder-only Transformer satisfying Assumption~\ref{def:transformer} can simulate a DTM (like $V$) running in $p'_V(n)$ steps by generating $p'_V(n)$ intermediate tokens. These tokens effectively form a Chain of Thought detailing the verifier's computation steps. These verification tokens are generated sequentially after the tokens representing $u$, forming a continuous part of the same AoT computational path. Let $k_V = p'_V(n)$ be the number of tokens for this verification phase.

\textit{3. Overall AoT Accepting Path and its Length:}
If $w \in L$, then there exists at least one certificate $u$ (of length $\le p_c(n)$) that $V$ accepts. The AoT-prompted Transformer $M_{AoT}$ is designed to explore the space of potential certificates. An accepting computational path for $M_{AoT}$ consists of:
\vspace{-3mm}
\begin{itemize}[leftmargin=4mm]
    \item A sequence of tokens representing the generation of a valid certificate $u$: requiring $k_u \le \mathcal{O}(p_c(n))$ tokens.
    \item A sequence of tokens representing the simulation of $V(\langle w, u \rangle)$ leading to acceptance by $V$: requiring $k_V = p'_V(n)$ tokens.
\end{itemize}
\vspace{-2mm}
The total number of intermediate tokens along this single accepting computational path is $a(n) = k_u + k_V$. This sum is bounded as follows:
\[ a(n) \le \mathcal{O}(p_c(n)) + p'_V(n). \]
Since $p_c(n)$ and $p'_V(n)$ are both polynomials in $n$, their sum $a(n)$ is also a polynomial in $n$. Note, that this upper bound applies to the accepting path specifically, not necessarily the full exploration (see remark~\ref{rem:aot-exp-exploration}).

If $V$ accepts $\langle w, u \rangle$, $M_{AoT}$ is instructed to output an accept signal. The AoT framework, as guided by its in-context algorithmic examples, allows the model to find such a path if one exists. If $w \notin L$, no such certificate exists, and consequently, no such accepting computational path of polynomial length can be completed by $M_{AoT}$ leading to acceptance; it will eventually halt and reject (as per the requirement that it decides $L$, which implies halting on all inputs after exploring relevant paths or exhausting search bounds defined by the AoT prompting strategy).

The definition of $\mathsf{AoT}(\mathtt{poly}(n))$ in Theorem~\ref{thm:AoT} specifically refers to the length $a(n)$ of this successful computational trace. The existence of such a polynomially-bounded accepting path for all $w \in L$ is guaranteed if the underlying $\mathsf{NP}$ problem has such certificates and verifiers.
Therefore, any language $L \in \mathsf{NP}$ can be decided by an AoT-prompted Transformer where an accepting computation involves generating $a(n)=\mathtt{poly}(n)$ intermediate tokens along that specific accepting path. Thus, $\mathsf{NP} \subseteq \mathsf{AoT}(\mathtt{poly}(n))$.

\textbf{Part II: $\mathsf{AoT}(\mathtt{poly}(n)) \subseteq \mathsf{NP}$}

This inclusion requires showing that if a language $L$ is decided by an AoT-prompted Transformer $M_{AoT}$ using at most $a(n)=\mathtt{poly}(n)$ intermediate tokens along any single accepting computational path, then $L \in \mathsf{NP}$.

To show $L \in \mathsf{NP}$, we must construct a polynomial-time deterministic verifier DTM, $V_{AoT}$, and demonstrate that for any $w \in L$, there exists a certificate $u_{cert}$ of polynomial length that $V_{AoT}$ can use to verify $w$'s membership in $L$ in polynomial time.

\textit{1. Certificate Definition:}
Let $w \in L$. Since $M_{AoT}$ decides $L$ and accepts $w$, by the definition of $\mathsf{AoT}(\mathtt{poly}(n))$ in Theorem~\ref{thm:AoT}, there exists at least one specific computational path (a sequence of generated intermediate thought tokens) that leads to $M_{AoT}$'s acceptance of $w$. Let this sequence of tokens be $S_{AoT} = (s_1, s_2, \dots, s_k)$, where $k \le a(n)$ and $a(n)$ is a polynomial in $n$. This sequence $S_{AoT}$ serves as our certificate, $u_{cert}$.
The length of this certificate, measured in the number of tokens, is $k$, which is polynomial in $n$. The bit length of $u_{cert}$ is $k \cdot P_{bits}$. With $P_{bits} = \mathcal{O}(\mathtt{log}(n+k))$ (from Assumption~\ref{def:transformer} on Logarithmic Precision for Parameters and Activations, applied to input $n$ and CoT/AoT length $k$), and since $k \le a(n) = \mathtt{poly}(n)$, $n+k$ is also polynomial in $n$. Thus, $\mathtt{log}(n+k)$ is polylogarithmic in $n$. The bit length of $u_{cert}$ is $k \cdot \mathcal{O}(\mathtt{log}(n+a(n)))$, which is $\mathtt{poly}(n) \cdot \mathtt{polylog}(n)$, and is therefore polynomial in $n$.

\textit{2. Verifier DTM ($V_{AoT}$) Construction:}
The verifier DTM $V_{AoT}$ takes as input the pair $\langle w, u_{cert} \rangle = \langle w, S_{AoT} \rangle$. $V_{AoT}$ must perform two main tasks in polynomial time:
\vspace{-3mm}
\begin{itemize}[leftmargin=4mm]
    \item Verify that $S_{AoT}$ is a valid computational path that $M_{AoT}$ could have legitimately generated starting from input $w$.
    \item Verify that this path $S_{AoT}$ results in $M_{AoT}$ accepting $w$.
\end{itemize}
\vspace{-2mm}
The DTM $V_{AoT}$ operates by simulating the AoT-prompted Transformer $M_{AoT}$ step-by-step, forced along the path dictated by the provided certificate $S_{AoT}$.
For each token $s_j$ in $S_{AoT}$ (for $j$ from $1$ to $k$):
\vspace{-3mm}
\begin{itemize}[leftmargin=4mm]
    \item $V_{AoT}$ considers the original input $w$ and the sequence of previously verified tokens from the certificate, $(s_1, \dots, s_{j-1})$, as the current prefix for $M_{AoT}$.
    \item $V_{AoT}$ then simulates a single forward pass of the Transformer $M_{AoT}$ (as defined by its architecture in Assumption~\ref{def:transformer}) to determine the token $s'_j$ that $M_{AoT}$ would deterministically generate given this prefix $(w, s_1, \dots, s_{j-1})$.
    \item $V_{AoT}$ compares this internally computed $s'_j$ with the token $s_j$ from the certificate. If $s'_j \neq s_j$, then $S_{AoT}$ is not a valid computational path that $M_{AoT}$ would have taken, so $V_{AoT}$ halts and rejects.
\end{itemize}

\textit{3. Time Complexity of $V_{AoT}$:}
The dominant operation within $V_{AoT}$ is the simulation of a single forward pass of the Transformer $M_{AoT}$ for each of the $k$ tokens in the certificate. Let $N_j = n+(j-1)$ be the length of the sequence (input $w$ + $j-1$ preceding thought tokens) upon which the $j$-th token is conditioned. The proof of Theorem 3 in~\cite{merrill2023expressive} ($\mathsf{CoT}(t(n)) \subseteq \widetilde{\mathsf{TIME}}(n^2 + t(n)^2)$) indicates that a single forward pass on a sequence of length $N_j$ can be simulated by a DTM in time polynomial in $N_j$, specifically $\mathcal{O}(N_j^2 \cdot \mathtt{polylog}(N_j))$, denoted $\widetilde{\mathcal{O}}(N_j^2)$.
Since $j \le k \le a(n)$, and $a(n)$ is polynomial in $n$, $N_j = n+j-1$ is also polynomial in $n$. Thus, the time to verify each token $s_j$ is $\widetilde{\mathcal{O}}((n+j-1)^2)$, which is polynomial in $n$.

The DTM $V_{AoT}$ performs this verification for all $k = |S_{AoT}|$ tokens. The total time for $V_{AoT}$ is the sum of times for each step:
\[ \text{Time}(V_{AoT}) = \sum_{j=1}^{k} \widetilde{\mathcal{O}}((n+j-1)^2). \]
Since $k \le a(n)$ and $a(n)$ is a polynomial in $n$ (let $a(n) = \mathcal{O}(n^{c_1})$ for some constant $c_1 \ge 0$), this sum is bounded by $k \cdot \widetilde{\mathcal{O}}((n+k)^2)$. Substituting $k = \mathcal{O}(n^{c_1})$:
\[ \text{Time}(V_{AoT}) = \mathcal{O}(n^{c_1}) \cdot \widetilde{\mathcal{O}}((n+\mathcal{O}(n^{c_1}))^2). \]
Let $c_{max} = \max(1, c_1)$. Then $n+\mathcal{O}(n^{c_1}) = \mathcal{O}(n^{c_{max}})$. The expression becomes:
\[ \text{Time}(V_{AoT}) = \mathcal{O}(n^{c_1}) \cdot \widetilde{\mathcal{O}}((n^{c_{max}})^2) = \mathcal{O}(n^{c_1}) \cdot \widetilde{\mathcal{O}}(n^{2c_{max}}) = \widetilde{\mathcal{O}}(n^{c_1 + 2c_{max}}). \]
This is a polynomial function of $n$ (as the $\widetilde{\mathcal{O}}$ notation hides polylogarithmic factors, which are absorbed into the overall polynomial nature).

\textit{4. Final Acceptance by $V_{AoT}$:}
If $V_{AoT}$ successfully verifies all $k$ tokens in $S_{AoT}$ (i.e., $s'_j = s_j$ for all $j=1, \dots, k$), it then needs to determine if this now-validated computational path $S_{AoT}$ indeed corresponds to an acceptance of $w$ by $M_{AoT}$. This typically involves checking the nature of the last generated token $s_k$ (e.g., if it's a special accept token) or by simulating the final classification layer of the Transformer after the full sequence $(w, S_{AoT})$ has been processed to see if it outputs an accept decision. This final check is part of the simulation of the $k$-th step (or an implicit $(k+1)$-th decision step based on the state after $s_k$) and is therefore completed within the polynomial time bound. If the path $S_{AoT}$ leads to an acceptance state for $M_{AoT}$, then $V_{AoT}$ accepts $\langle w, S_{AoT} \rangle$. Otherwise (e.g., if the path does not end in acceptance, or if any $s_j$ was inconsistent during verification), $V_{AoT}$ rejects.

Since the certificate $S_{AoT}$ has polynomial length (in terms of bits, as derived from $k$ tokens and logarithmic precision), and the deterministic verifier DTM $V_{AoT}$ verifies it in polynomial time with respect to $n$, it follows by the definition of $\mathsf{NP}$ that the language $L$ decided by $M_{AoT}$ is in $\mathsf{NP}$.
Therefore, $\mathsf{AoT}(\mathtt{poly}(n)) \subseteq \mathsf{NP}$.

Combining the inclusions from \textbf{Part I} ($\mathsf{NP} \subseteq \mathsf{AoT}(\mathtt{poly}(n))$) and \textbf{Part II} ($\mathsf{AoT}(\mathtt{poly}(n)) \subseteq \mathsf{NP}$), we conclude that \( \mathsf{AoT}(\mathtt{poly}(n)) = \mathsf{NP}. \)

\end{proof}

\subsection{Proof of Theorem~\ref{thm:CoT_exp}: \emph{$\mathsf{CoT}(\mathtt{exp}(n)) = \mathsf{EXP}$}}\label{subsec:proof-CoT_exp}

\begin{proof}

The notation $\mathtt{exp}(n)$ signifies a function $t(n)$ such that $t(n) = O(2^{p(n)})$, where $p(n)$ is a polynomial in $n$. The complexity class $\mathsf{EXP}$ is defined according to Definition~\ref{def:time_class} as \( \mathsf{EXP} = \bigcup_{k \in \mathbb{N}} \mathsf{TIME}(2^{n^k}). \)
The proof is demonstrated in two parts.

\textbf{Part I: $\mathsf{EXP} \subseteq \mathsf{CoT}(\mathtt{exp}(n))$}

This inclusion is established by demonstrating that a CoT-augmented Transformer, under the specified architectural assumptions, can simulate any DTM that operates within an exponential time bound.

Let $L$ be an arbitrary language in the class $\mathsf{EXP}$. By the definition of $\mathsf{EXP}$, there exists a DTM $M_L$ and a non-negative integer constant $k_0$ such that $M_L$ decides $L$, and for any input $w$ of length $n = \lvert w \rvert$, $M_L$ halts in $T_{M_L}(n) = O(2^{n^{k_0}})$ time. Let $p_L(n) = n^{k_0}$; thus, $M_L$ runs in $O(2^{p_L(n)})$ time. This means $L \in \mathsf{TIME}(O(2^{p_L(n)}))$.

We leverage Theorem 2 in~\cite{merrill2023expressive} for $L \in \mathsf{EXP}$, and set the DTM running time $t(n)$ to correspond to $T_{M_L}(n)$. If $T_{M_L}(n) \le c \cdot 2^{p_L(n)}$ for some constant $c$, we can choose the number of CoT steps to be $t_{\text{CoT}}(n) = 2^{p'_L(n)}$ where $p'_L(n)$ is a polynomial such that $c \cdot 2^{p_L(n)} \le 2^{p'_L(n)}$ for sufficiently large $n$ (e.g., $p'_L(n) = p_L(n) + \lceil \mathtt{log}_2 c \rceil$). This $t_{\text{CoT}}(n)$ is an $\mathtt{exp}(n)$ function. The Transformer will thus generate $t_{\text{CoT}}(n)$ CoT tokens to simulate $M_L$.

The architectural assumptions of Logarithmic Embedding Dimension ($P_{\text{bits}}(W,A) = \mathcal{O}(\mathtt{log}(n + t(n)))$ from Assumption~\ref{def:transformer}) must be compatible. With $t(n) = t_{\text{CoT}}(n) = \mathcal{O}(2^{p'_L(n)})$, the precision becomes $P_{\text{bits}} = \mathcal{O}(\mathtt{log}(n + 2^{p'_L(n)}))$. For large $n$, $2^{p'_L(n)}$ dominates, leading to:
\[ P_{\text{bits}} = \mathcal{O}(\mathtt{log}(2^{p'_L(n)})) = \mathcal{O}(p'_L(n)). \]
This polynomial precision (in $n$) is consistent with the $\mathcal{O}(\mathtt{log}(T+N))$ precision requirement for their DTM simulation construction (where $T$ is DTM steps, $N$ is input length). This is a crucial point: the logarithmic precision is relative to the total sequence length; if the CoT is exponential, the required bit precision becomes polynomial in $n$. The Transformer simulates $M_L$ using $t_{\text{CoT}}(n)$ CoT steps.

Next, by Corollary 2.1 in~\cite{merrill2023expressive} \( \mathsf{TIME}(t(n)) \subseteq \mathsf{CoT}(t(n)). \)
Since $L \in \mathsf{TIME}(T_{M_L}(n))$ and $T_{M_L}(n) \le t_{\text{CoT}}(n)$ where $t_{\text{CoT}}(n)$ is $\mathtt{exp}(n)$, by applying Corollary 2.1, we have $L \in \mathsf{CoT}(t_{\text{CoT}}(n))$.
Therefore, $L \in \text{CoT}(\mathtt{exp}(n))$.
As $L$ was an arbitrary language in $\mathsf{EXP}$, this establishes the inclusion $\mathsf{EXP} \subseteq \mathsf{CoT}(\mathtt{exp}(n))$.

\textbf{Part II: $\mathsf{CoT}(\mathtt{exp}(n)) \subseteq \mathsf{EXP}$}

This inclusion is demonstrated by showing that the computation performed by a CoT-augmented Transformer using $\mathtt{exp}(n)$ CoT steps can be simulated by a DTM in exponential time.

Let $L$ be an arbitrary language in $\mathsf{CoT}(\mathtt{exp}(n))$. By definition, there exists a CoT-augmented Transformer, $M_{\text{CoT}}$ (satisfying Assumption~\ref{def:transformer}), that decides $L$ by generating $t_{\text{CoT}}(n)$ intermediate CoT tokens, where $t_{\text{CoT}}(n) = \mathcal{O}(2^{p_M(n)})$ for some polynomial $p_M(n)$.

According to Theorem 3 in~\cite{merrill2023expressive}, the relationship is \( \text{CoT}(t(n)) \subseteq \widetilde{\mathsf{TIME}}(n^2 + t(n)^2). \)
The $\widetilde{\mathsf{TIME}}$ notation hides polylogarithmic factors of its argument, i.e., the actual DTM simulation time is $\mathcal{O}((n^2 + t(n)^2) \cdot (\mathtt{log}(n + t(n)))^j)$ for some constant $j$.

Substitute $t_{\text{CoT}}(n) = \mathcal{O}(2^{p_M(n)})$ into this time bound. The term $(t_{\text{CoT}}(n))^2$ becomes:
\[ (t_{\text{CoT}}(n))^2 = (\mathcal{O}(2^{p_M(n)}))^2 = \mathcal{O}(2^{2p_M(n)}). \]
Let $p'_M(n) = 2p_M(n)$, which is also a polynomial. So, $(t_{\text{CoT}}(n))^2 = \mathcal{O}(2^{p'_M(n)})$.
The base complexity $n^2 + (t_{\text{CoT}}(n))^2 = n^2 + \mathcal{O}(2^{p'_M(n)})$. For large $n$ (assuming $p'_M(n)$ is not identically zero), this is dominated by $\mathcal{O}(2^{p'_M(n)})$.

Now, consider the polylogarithmic factor $(\mathtt{log}(n + t_{\text{CoT}}(n)))^j$. Since $t_{\text{CoT}}(n) = \mathcal{O}(2^{p_M(n)})$, for large $n$, $n + t_{\text{CoT}}(n)$ is $\mathcal{O}(2^{p_M(n)})$. So, $\mathtt{log}(n + t_{\text{CoT}}(n)) = \mathtt{log}(\mathcal{O}(2^{p_M(n)})) = \mathcal{O}(\mathtt{log}(2^{p_M(n)})) = \mathcal{O}(p_M(n))$. The polylogarithmic factor is $(\mathcal{O}(p_M(n)))^j$, which is a polynomial in $n$. Let this polynomial be $P_{\text{poly\_log}}(n)$.

The total DTM simulation time is $\mathcal{O}(2^{p'_M(n)} \cdot P_{\text{poly\_log}}(n))$. To show this is an exponential time bound of the form $2^{\text{poly}(n)}$, we rewrite $P_{\text{poly\_log}}(n)$ as $2^{\mathtt{log}_2(P_{\text{poly\_log}}(n))}$. The time bound then becomes:
\[ \mathcal{O}(2^{p'_M(n)} \cdot 2^{\mathtt{log}_2(P_{\text{poly\_log}}(n))}) = \mathcal{O}(2^{p'_M(n) + \mathtt{log}_2(P_{\text{poly\_log}}(n))}). \]
Let $p''_M(n) = p'_M(n) + \mathtt{log}_2(P_{\text{poly\_log}}(n))$.
Since $p'_M(n)$ is a polynomial in $n$, and $P_{\text{poly\_log}}(n)$ is also a polynomial in $n$, $\mathtt{log}_2(P_{\text{poly\_log}}(n))$ grows slower than any polynomial of positive degree (e.g., if $P_{\text{poly\_log}}(n) = n^a$, then $\mathtt{log}_2(n^a) = a \mathtt{log}_2 n$).
\begin{itemize}[leftmargin = 4mm]
    \item If $p_M(n)$ (and thus $p'_M(n)$) is a polynomial of degree $d \ge 1$, then $p'_M(n)$ dominates $\mathtt{log}_2(P_{\text{poly\_log}}(n))$. Thus, $p''_M(n) = \mathcal{O}(p'_M(n))$, which is polynomial. The DTM runs in $\mathcal{O}(2^{\text{poly}(n)})$ time.
    \item If $p_M(n)$ is a constant (degree 0), say $c_0$, then $t_{\text{CoT}}(n) = \mathcal{O}(2^{c_0}) = \mathcal{O}(1)$. In this case, the original simulation time is $\widetilde{\mathcal{O}}(n^2) = \mathcal{O}(n^2 \cdot (\mathtt{log} n)^j)$, which is polynomial time. Since $\mathsf{P} \subseteq \mathsf{EXP}$, this scenario is consistent with $p''_M(n)$ effectively becoming a function that leads to an overall polynomial time when the base $n^2$ term dominates.
\end{itemize}
More generally, for $t_{\text{CoT}}(n) = \mathcal{O}(2^{p_M(n)})$, the exponent $p''_M(n)$ is bounded by some polynomial in $n$. The DTM simulates $M_{\text{CoT}}$ in time $\mathcal{O}(2^{p''_M(n)})$. By definition of $\mathsf{EXP}$ (time $2^{\text{poly}(n)}$), this signifies that $L \in \mathsf{EXP}$. As $L$ was an arbitrary language in $\mathsf{CoT}(\mathtt{exp}(n))$, we conclude that $\mathsf{CoT}(\mathtt{exp}(n)) \subseteq \mathsf{EXP}$.

By combining the inclusions from \textbf{Part I} ($\mathsf{EXP} \subseteq \mathsf{CoT}(\mathtt{exp}(n))$) and \textbf{Part II} ($\mathsf{CoT}(\mathtt{exp}(n)) \subseteq \mathsf{EXP}$), it directly follows that \( \mathsf{CoT}(\mathtt{exp}(n)) = \mathsf{EXP}. \)

This equivalence demonstrates that decoder-only Transformers (under Assumption~\ref{def:transformer}) augmented with a number of CoT tokens that is exponential in the input length possess computational power equivalent to that of exponential-time Deterministic Turing Machines.

\end{proof}

\subsection{Proof of Theorem~\ref{thm:AoT_exp}: \emph{$\mathsf{AoT}(\mathtt{exp}(n)) = \mathsf{NEXP}$}}\label{subsec:proof-AoT_exp}

\begin{proof}
The complexity class $\mathsf{NEXP}$ is defined as:
\( \mathsf{NEXP} = \bigcup_{k \in \mathbb{N}} \mathsf{NTIME}(2^{n^k}). \) An exponential function $a(n) = \mathtt{exp}(n)$ means $a(n) = O(2^{p_A(n)})$ for some polynomial $p_A(n)$. The proof consists of two parts: $\mathsf{NEXP} \subseteq \mathsf{AoT}(\mathtt{exp}(n))$ and $\mathsf{AoT}(\mathtt{exp}(n)) \subseteq \mathsf{NEXP}$. We rely on the definitions provided in the Preliminaries~\ref{app:theory} and the architectural assumptions in Assumption~\ref{def:transformer}.

\textbf{Part I: $\mathsf{NEXP} \subseteq \mathsf{AoT}(\mathtt{exp}(n))$}

This part demonstrates that any language $L \in \mathsf{NEXP}$ can be decided by an AoT-prompted Transformer using a number of intermediate tokens that is exponential in the input length $n$ along any accepting path.

Let $L$ be an arbitrary language in $\mathsf{NEXP}$. By definition, $L \in \mathsf{NTIME}(T(n))$ for $T(n) = 2^{p_L(n)}$ for some polynomial $p_L(n)$. As stated in the Definition of Deterministic and Nondeterministic Time Classes~\ref{def:time_class}, membership in $\mathsf{NTIME}(t(n))$ implies there exists a deterministic Turing machine (DTM) $V$ (the verifier) such that for every input string $w$ of length $n = |w|$:
\begin{gather*}
  w \in L \iff \exists u \in \Sigma^* \text{ (certificate/witness) s.t. } \\ 
  |u| = O(2^{p_L(n)}), \text{ and } V \text{ accepts } \langle w, u \rangle \text{ in } O(2^{p_L(n)}) \text{ time}.
\end{gather*}
Let $p'(n)$ be a polynomial (potentially $p_L(n)$ adjusted by constants) such that $|u| \le c_1 \cdot 2^{p'(n)}$ and $V$ runs in at most $T_V(n) = c_2 \cdot 2^{p'(n)}$ steps for constants $c_1, c_2$.

We construct an AoT-prompted Transformer, denoted $M_{AoT}$, satisfying Assumption~\ref{def:transformer}, that decides $L$. On input $w$, the AoT process for $M_{AoT}$ proceeds as follows:

\textit{1. Certificate Generation (Simulating Nondeterministic Guess via AoT Exploration):}
The AoT prompting supplies $M_{AoT}$ with examples demonstrating algorithmic search or construction. This capability is leveraged to explore the space of potential certificates and generate a candidate $u$. If $w \in L$, a valid certificate $u$ of length $O(2^{p'(n)})$ exists and, by Definition~\ref{def:aot-complexity}, the search space is sufficiently large to include all such candidates, so in particular $u$. $M_{AoT}$ generates a sequence of intermediate tokens representing this candidate $u$. Assuming each symbol of $u$ is encoded by a constant (or polylogarithmic) number of tokens, the number of tokens required to represent $u$, denoted $k_u$, is $k_u = O(|u|) = O(2^{p'(n)})$. This is an $\mathtt{exp}(n)$ number of tokens.

\textit{2. Verification (Simulating Verifier DTM $V$):}
After generating the tokens for a candidate $u$, $M_{AoT}$ simulates the execution of the DTM verifier $V$ on $\langle w, u \rangle$. The verifier $V$ runs in $T_V(n) = O(2^{p'(n)})$ time. Based on the established capability of Transformers under Assumption~\ref{def:transformer} to simulate DTMs (using the logic from Theorem 2 in~\cite{merrill2023expressive}), $M_{AoT}$ can simulate the $T_V(n)$ steps of $V$ by generating $k_V = O(T_V(n)) = O(2^{p'(n)})$ intermediate tokens. This simulation is feasible because the required precision $P_{bits} = O(\mathtt{log}(n + k_u + k_V)) = O(\mathtt{log}(n + 2^{p'(n)}))$, which simplifies for large $n$ to:
\[ P_{bits} = O(p'(n)). \]
This polynomial precision (in $n$) is consistent with Assumption~\ref{def:transformer}.

\textit{3. Overall AoT Accepting Path and its Length:}
If $w \in L$, there exists at least one certificate $u$ for which $V(\langle w, u \rangle)$ accepts. The AoT prompting guides $M_{AoT}$ to find such a computational path. This accepting path comprises:
\begin{itemize}[leftmargin=4mm]
  \item $k_u = O(2^{p'(n)})$ tokens for generating $u$.
  \item $k_V = O(2^{p'(n)})$ tokens for simulating $V$'s accepting computation.
\end{itemize}
The total number of intermediate tokens along this single accepting path is $a_{\text{path}}(n) = k_u + k_V = O(2^{p'(n)}) + O(2^{p'(n)}) = O(2^{p'(n)})$. Since $p'(n)$ is a polynomial, $a_{\text{path}}(n)$ is an $\mathtt{exp}(n)$ function. This path length must be bounded by the overall AoT capacity $a(n) = O(2^{p_A(n)})$, which holds if we choose $p_A(n)$ appropriately (e.g., $p_A(n) = p'(n)$). Note, that this upper bound applies to the accepting path specifically, not necessarily the full exploration (see Remark~\ref{rem:aot-exp-exploration}).

If $w \notin L$, no certificate $u$ exists that $V$ accepts. Consequently, no AoT path corresponding to guessing and successfully verifying a certificate will lead to acceptance. $M_{AoT}$ halts and rejects (as it decides $L$).

The existence of such an exponentially-bounded accepting path for all $w \in L$ (and no such path for $w \notin L$) means $L \in \mathsf{AoT}(\mathtt{exp}(n))$. Therefore, $\mathsf{NEXP} \subseteq \mathsf{AoT}(\mathtt{exp}(n))$.

\textbf{Part II: $\mathsf{AoT}(\mathtt{exp}(n)) \subseteq \mathsf{NEXP}$}

This part shows that if a language $L$ is decided by an AoT-prompted Transformer $M_{AoT}$ using $a(n) = \mathtt{exp}(n)$ tokens along any accepting path ($a(n) = O(2^{p_A(n)})$ for polynomial $p_A(n)$), then $L \in \mathsf{NEXP}$.

To prove $L \in \mathsf{NEXP}$, we construct a NEXP verifier for $L$. We define a certificate $u_{\text{cert}}$ of length $\mathtt{exp}(n)$ and a DTM verifier $V_{AoT}$ that runs in $\mathtt{exp}(n)$ time, such that $V_{AoT}$ accepts $\langle w, u_{\text{cert}} \rangle$ if and only if $w \in L$.

\textit{1. Certificate Definition:}
If $w \in L$, by definition of $M_{AoT}$ deciding $L$ in $\text{AoT}(\mathtt{exp}(n))$, there exists at least one accepting sequence of intermediate AoT tokens $S_{\text{AoT}} = (s_1, s_2, \dots, s_k)$, where $k \le a(n) = O(2^{p_A(n)})$. This sequence $S_{\text{AoT}}$ is our certificate, $u_{\text{cert}}$. The number of tokens is $k = O(2^{p_A(n)})$.
The bit length requires precision $P_{bits}$. From Assumption~\ref{def:transformer}, $P_{bits} = O(\mathtt{log}(n+k))$. Since $k = O(2^{p_A(n)})$, $P_{bits} = O(\mathtt{log}(n + 2^{p_A(n)})) = O(p_A(n))$ (for polynomial $p_A(n)$). The total bit length of $u_{\text{cert}}$ is $|u_{\text{cert}}| = k \cdot P_{bits}$, which evaluates to:
\[ |u_{\text{cert}}| = O(2^{p_A(n)}) \cdot O(p_A(n)) = O(2^{p_A(n) + \mathtt{log}_2(p_A(n))}). \]
Let $p'_A(n) = p_A(n) + \mathtt{log}_2(p_A(n))$. Since $p_A(n)$ is polynomial, $p'_A(n)$ is also polynomial (as $\mathtt{log}_2$ of a polynomial grows slower than the polynomial). Thus, the bit length $|u_{\text{cert}}| = O(2^{p'_A(n)})$, which is $\mathtt{exp}(n)$.

\textit{2. Verifier DTM ($V_{AoT}$) Construction:}
The DTM $V_{AoT}$ takes $\langle w, u_{\text{cert}} \rangle = \langle w, S_{\text{AoT}} \rangle$ as input. It deterministically simulates $M_{AoT}$ step-by-step, verifying against $S_{\text{AoT}}$:
For $j = 1$ to $k$:
\vspace{-2mm}
\begin{itemize}[leftmargin=4mm]
  \item Consider the prefix $(w, s_1, \dots, s_{j-1})$.
  \item Simulate one forward pass of $M_{AoT}$ on this prefix to compute the next token $s'_j$.
  \item Compare $s'_j$ with $s_j$. If $s'_j \neq s_j$, reject.
\end{itemize}
\vspace{-1mm}
If all $k$ tokens match, check if the path $S_{\text{AoT}}$ corresponds to an acceptance state in $M_{AoT}$. If yes, accept $\langle w, S_{\text{AoT}} \rangle$. Otherwise, reject.

\textit{3. Time Complexity of $V_{AoT}$:}
The core cost is simulating $k$ forward passes. Let $N_j = n + (j-1)$ be the sequence length at step $j$. From the analysis supporting Theorem 3 in~\cite{merrill2023expressive}, the simulation time for one step, $T_{\text{step}}(N_j)$, is $O(N_j^2 \cdot \text{poly}(P_{\text{bits}}))$. Here, $P_{\text{bits}} = O(p_A(n))$, so let $P_{\text{log}}(n) = \text{poly}(p_A(n))$, a polynomial in $n$. Thus, $T_{\text{step}}(N_j) = O(N_j^2 \cdot P_{\text{log}}(n))$.
The total time for $V_{AoT}$ is $\sum_{j=1}^{k} T_{\text{step}}(N_j)$. This sum is bounded by $k \times T_{\text{step}}(N_k)$, where $N_k = n + k - 1$. This leads to:
\[ \text{Time}(V_{AoT}) \le k \cdot O(N_k^2 \cdot P_{\text{log}}(n)). \]
Substituting $k = O(2^{p_A(n)})$ and $N_k = O(n + 2^{p_A(n)}) = O(2^{p_A(n)})$ (for large $n$), we get:
\[ \text{Time}(V_{AoT}) \le O(2^{p_A(n)}) \cdot O((O(2^{p_A(n)}))^2 \cdot P_{\text{log}}(n)) = O(2^{3p_A(n)} \cdot P_{\text{log}}(n)). \]
Since $P_{\text{log}}(n)$ is polynomial, we write $P_{\text{log}}(n) = 2^{\mathtt{log}_2(P_{\text{log}}(n))}$. The time complexity becomes:
\[ \text{Time}(V_{AoT}) \le O(2^{3p_A(n) + \mathtt{log}_2(P_{\text{log}}(n))}). \]
Let $p''_A(n) = 3p_A(n) + \mathtt{log}_2(P_{\text{log}}(n))$. As $p_A(n)$ and $P_{\text{log}}(n)$ are polynomials, $p''_A(n)$ is also a polynomial in $n$. Therefore, the time complexity is $\text{Time}(V_{AoT}) = O(2^{p''_A(n)})$, which is $\mathtt{exp}(n)$.

We have constructed a DTM verifier $V_{AoT}$ that takes an input $w$ and a certificate $u_{\text{cert}} = S_{\text{AoT}}$ of length $|u_{\text{cert}}| = O(2^{p'_A(n)}) = \mathtt{exp}(n)$, and verifies in time $\text{Time}(V_{AoT}) = O(2^{p''_A(n)}) = \mathtt{exp}(n)$ whether $w \in L$. This precisely matches the verifier definition of $\mathsf{NEXP}$.
Therefore, $\mathsf{AoT}(\mathtt{exp}(n)) \subseteq \mathsf{NEXP}$.

Combining the inclusions from \textbf{Part I} ($\mathsf{NEXP} \subseteq \mathsf{AoT}(\mathtt{exp}(n))$) and \textbf{Part II} ($\mathsf{AoT}(\mathtt{exp}(n)) \subseteq \mathsf{NEXP}$), we conclude that \( \mathsf{AoT}(\mathtt{exp}(n)) = \mathsf{NEXP}. \)
This establishes that decoder-only Transformers, under Assumption~\ref{def:transformer}, when augmented with an AoT generating an exponential number of tokens along any single accepting path, possess computational power equivalent to that of exponential-time Nondeterministic Turing Machines.

\end{proof}

\subsection{Proof of Theorem~\ref{thm:core_reasoning_tokens}: \emph{Core Reasoning Tokens}}\label{subsec:proof-core_reasoning_tokens}
\begin{proof}
Let $M_{T}$ be a decoder-only Transformer (satisfying Assumption~\ref{def:transformer}) that decides a language $L$ by implicitly running an internal algorithm $A_{M_{T}}$ through its generation of intermediate tokens. We define the \emph{core computational trace} $S_{\text{core}}$ (length $k_{\text{core}}(n)$) to be the shortest subsequence of tokens capturing all essential, progressive computational steps of $A_{M_{T}}$ (redundant tokens, self-corrections, and filler contents are not included). Evidently~\citep{zeng2025revisiting, li2025sos1, wu2025more, ballon2025relationship, su2025between}, this core length is bounded by the total:
\[
k_{\text{core}}(n) \le k_{\text{total}}(n),
\]
where $k_{\text{total}}(n)$ is the total number of tokens produced by $M_{T}$ on inputs of size $n$.

The complexity class governing $L$ depends jointly on the growth of $k_{\text{core}}(n)$ and on the nature of $A_{M_{T}}$:
\begin{itemize}[leftmargin=4mm]
  \item Deterministic (CoT case). If $A_{M_{T}}$ operates as a step-by-step deterministic computation (analogous to a DTM), then:
  \[
    k_{\text{core}}(n)=
    \begin{cases}
      O\bigl(\mathtt{poly}(n)\bigr) &\implies L\in\mathsf{P},\\[4pt]
      O\bigl(\mathtt{exp}(n)\bigr)  &\implies L\in\mathsf{EXP}.
    \end{cases}
  \]
  \item Nondeterministic (AoT case). If an accepting computation of $A_{M_{T}}$ amounts to guessing and then verifying a witness (analogous to a NTM), then:
  \[
    k_{\text{core}}(n)=
    \begin{cases}
      O\bigl(\mathtt{poly}(n)\bigr) &\implies L\in\mathsf{NP},\\[4pt]
      O\bigl(\mathtt{exp}(n)\bigr)  &\implies L\in\mathsf{NEXP}.
    \end{cases}
  \]
\end{itemize}

Let $C_{poly}(A_{M_{T}})$ denote the polynomial complexity class and $C_{exp}(A_{M_{T}})$ the exponential complexity class that corresponds to the computational nature of $A_{M_{T}}$, defined as:
\[
  C_{poly}(A_{M_{T}}) =
    \begin{cases}
      \mathsf{P}  &\text{if } A_{M_{T}} \text{ is deterministic},\\
      \mathsf{NP} &\text{if } A_{M_{T}} \text{ is nondeterministic},
    \end{cases}
\]
\[
  C_{exp}(A_{M_{T}}) =
    \begin{cases}
      \mathsf{EXP}  &\text{if } A_{M_{T}} \text{ is deterministic},\\
      \mathsf{NEXP} &\text{if } A_{M_{T}} \text{ is nondeterministic}.
    \end{cases}
\]
The following hold:
\begin{enumerate}[label=(\arabic*), leftmargin=8mm]
  \item If $k_{\text{core}}(n)=O\left(\mathtt{poly}(n)\right)$, then $L$ lies in $C_{poly}\bigl(A_{M_{T}}\bigr)$, regardless of any super-polynomial overhead in $k_{\text{total}}(n)$.
  \item To decide a language $L\in C_{exp}\bigl(A_{M_{T}}\bigr)\setminus C_{poly}\bigl(A_{M_{T}}\bigr)$, the algorithm must exhibit $k_{\text{core}}(n)=O\left(\mathtt{exp}(n)\right)$; mere exponential length of $k_{\text{total}}(n)$ does not suffice if $k_{\text{core}}(n)$ remains polynomial.
\end{enumerate}

The key idea is that computational power is governed by the minimal, redundancy-free trace $S_{\text{core}}$'s length $k_{\text{core}}(n)$, not the possibly much larger $k_{\text{total}}(n)$. Imagine an idealized Transformer $M^{*}_{T}$ that outputs exactly $S_{\text{core}}$ and nothing more. Because $M^{*}_{T}$ runs the same algorithm $A_{M_{T}}$, its complexity is governed by $k_{\text{core}}(n)$ and by whether $A_{M_{T}}$ is deterministic or nondeterministic.

\noindent\textbf{Part I:}\;Suppose $k_{\text{core}}(n)=O\bigl(\mathtt{poly}(n)\bigr)$. Two cases:
\begin{itemize}[leftmargin=4mm]
  \item \emph{Deterministic (CoT case).} Then $M^{*}_{T}$ performs a polynomial-length simulation of a DTM, so by Theorem~\ref{thm:CoT} we have $L\in\mathsf{P}$.
  \item \emph{Nondeterministic (AoT case).} Here $M^{*}_{T}$ outputs a polynomial-size witness–verification pair; applying Theorem~\ref{thm:AoT} yields $L\in\mathsf{NP}$.
\end{itemize}
Either way, $L\in C_{poly}\bigl(A_{M_{T}}\bigr)$. The extra $k_{\text{total}}(n)-k_{\text{core}}(n)$ tokens never raise the underlying complexity, even though simulating the raw output stream could take exponential time.

\medskip
\noindent\textbf{Part II:}\;Assume $M_{T}$ decides a language $L\in C_{exp}\bigl(A_{M_{T}}\bigr)\setminus C_{poly}\bigl(A_{M_{T}}\bigr)$. If, contrary to claim, $k_{\text{core}}(n)=O\bigl(\mathtt{poly}(n)\bigr)$, Part I would force $L$ into $C_{poly}\bigl(A_{M_{T}}\bigr)$ as a contradiction. Hence $k_{\text{core}}(n)$ is super-polynomial.

Moreover, the constructive inclusions from Theorems~\ref{thm:CoT_exp} and~\ref{thm:AoT_exp} show that an exponential-length core computational trace is both necessary and sufficient for languages that provably separate $\mathsf{P}$ from $\mathsf{EXP}$ or $\mathsf{NP}$ from $\mathsf{NEXP}$. Therefore $k_{\text{core}}(n)=O \bigl(\mathtt{exp}(n)\bigr)$ is required. In conclusion, only by scaling the effective part of its reasoning tokens can a Transformer transcend polynomial-time capabilities.
\end{proof}

\end{document}